\documentclass[11pt]{article}

\usepackage[preprint]{acl}

\usepackage{times}
\usepackage{latexsym}

\usepackage[T1]{fontenc}

\usepackage[utf8]{inputenc}

\usepackage{microtype}

\usepackage{inconsolata}

\usepackage{graphicx}
\usepackage{booktabs}
\usepackage{multirow}
\usepackage{listings}
\usepackage{colortbl}
\usepackage{xcolor}
\usepackage{booktabs}
\usepackage{array}
\usepackage{enumitem}
\usepackage{amsmath}
\usepackage{algorithm}
\usepackage{algpseudocode}
\usepackage{tikz}
\usepackage{pgfplots}
\usepackage{tabularx}
\usepackage{xurl}
\usetikzlibrary{intersections}
\usepackage{pifont}
\usepackage{makecell}

\usepackage{adjustbox}

\pgfplotsset{compat=1.18}

\definecolor{morandiblue}{RGB}{104,139,171}
\definecolor{morandirose}{RGB}{181,133,137}
\definecolor{morandigreen}{RGB}{123,151,137}
\definecolor{morandigray}{RGB}{145,142,137}

\newcommand{\cmark}{\textcolor{teal}{\ding{51}}}  
\newcommand{\xmark}{\ding{55}}
\usepgfplotslibrary{fillbetween}

\lstdefinestyle{promptstyle}{
  basicstyle=\ttfamily\scriptsize,
  breaklines=true,
  breakatwhitespace=false,
  columns=fullflexible,
  keepspaces=true,
  showstringspaces=false,
  frame=none
}

\title{\textsc{MemRole}: From Memories to Decisions for Long-Horizon Reasoning in Role-Playing Agents}

\author{
 \textbf{Mengting Li}\textsuperscript{$1,*$},
 \textbf{Xilong Cheng}\textsuperscript{$2,*$},
 \textbf{Zhihao Xiao}\textsuperscript{$1,*$},
 \textbf{Zhenhua Xu}\textsuperscript{3},
\\
 \textbf{Jingyuan Zhou}\textsuperscript{4},
 \textbf{Linfeng Li}\textsuperscript{1},
 \textbf{Long Zhang}\textsuperscript{5},
 \textbf{Borui Cai}\textsuperscript{$1,\dagger$},
 \textbf{Yunxiao Qin}\textsuperscript{$2,\dagger$},
\\[4pt]
 \textsuperscript{1}Beihang University,
 \textsuperscript{2}Communication University of China,
\\
 \textsuperscript{3}Zhejiang University,
 \textsuperscript{4}Guangxi University,
 \textsuperscript{5}Kuaishou Technology
\\[4pt]
 \small{
   \href{mailto:li.mengting@buaa.edu.cn}
   {li.mengting@buaa.edu.cn};
   \href{mialto:chengzhengyu330@cuc.edu.cn,}
   {chengzhengyu330@cuc.edu.cn}
 }
}

\RequirePackage{fix-cm}
\begin{document}
\maketitle

\begingroup
\renewcommand{\thefootnote}{}
\footnotetext[0]{
$*$ Equal contribution.
\quad
$\dagger$ Corresponding author.
\\
Code, model checkpoints, and datasets will be made publicly available promptly upon acceptance.
}
\endgroup
\begin{abstract}
LLM-based role-playing agents are fluent at imitating a character's style, yet struggle to make role-consistent decisions over long narrative horizons. Such decisions are shaped by experiences spanning many prior scenes, which are often partial and may point in different directions. Existing methods inject cross-scene memories as context, but do not reason about how past events shape a character's current motivations and behavior. We propose \textsc{MemRole}, a framework that makes this memory-to-decision process explicit. It introduces \textbf{L}ong-\textbf{H}orizon \textbf{M}emory \textbf{T}hinking (\textbf{LHMT}), a reasoning schema that decomposes hidden reasoning into context understanding, memory activation, and response strategy planning, trained on a novel-derived dataset whose character-grounded, timeline-constrained memory pools limit each character to what it could plausibly know at the current narrative point. Beyond supervised fine-tuning, we propose \textbf{S}elf-\textbf{R}efining \textbf{O}n-\textbf{P}olicy \textbf{S}elf-\textbf{D}istillation (\textbf{SR-OPSD}): the model critiques its own rollouts against the reference under a data-driven rubric, and a critique-conditioned teacher view of the same model distills token-level guidance back into the policy. \textsc{MemRole}-32B improves over its backbone by 15.2 points on memory-grounded role-playing, performs on par with frontier LLMs such as GPT-5.4, and achieves the best character-motivation recognition accuracy on CroSS-MR.
\end{abstract}

\section{Introduction}

\begin{figure}[t]
  \centering
  \includegraphics[width=\linewidth]{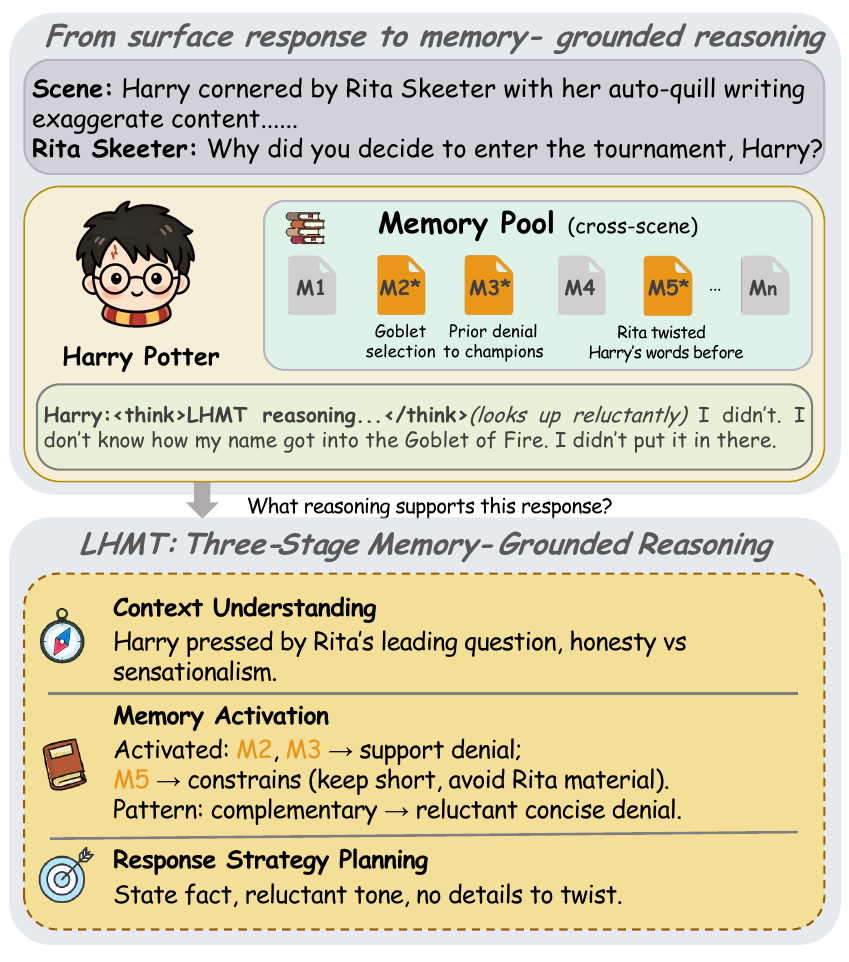}
  \caption{Illustration of Long-Horizon Memory Thinking (LHMT), which transforms cross-scene memories into a role-consistent response through three structured stages.}
  \label{fig:memory_reasoning}
\end{figure}

The recent progress of role-playing agents (RPAs) has benefited substantially from the rapid development of large language models (LLMs), enabling them to emulate a character’s speaking style, personality traits, and background knowledge with increasing fluency~\citep{shao2023character, zhou2024characterglm, wang2025coser}.
Yet fluency is not fidelity. What ultimately distinguishes a character is not how they talk in a single turn, but the decisions they make—and a character's decisions are rarely determined by the immediate context alone~\cite{mcadams2011narrative,conway2000construction}.
They are shaped by experiences that span many prior scenes: past humiliations, broken promises, shifting loyalties. 
A choice that looks abrupt or even shocking in the moment is in fact driven by this long-horizon memory, which renders it coherent and, in hindsight, even inevitable.

We refer to this ability as \textbf{long-horizon reasoning} in role-playing: reasoning over how memories spanning many prior scenes, often partial and not always mutually reinforcing, drive a character's decisions in the present.
At consequential moments, such memory shapes not only what a character decides, the externally observable action or stance, but also the motivations and emotions that, though unobservable, are carried into that decision~\citep{pillemer2003directive, graesser1994constructing}.
Yet most existing methods supply cross-scene memory through retrieval-augmented generation, surfacing relevant past snippets as supplementary background to reinforce consistency rather than reasoning about how those events drive the current decision~\citep{wang2024rolellm, ahn-etal-2024-timechara, cheng2026psymem, zhang2026psymembehav, 10.1145/3770855.3818027}. 
Approaches that go further still fall short: some tie memory to the immediate turn~\citep{liu2025cogdual}, while others, closest to our setting, reduce role-playing to a multiple-choice benchmark that scores a character's choice but leaves the underlying motivations and emotions unexamined~\citep{xu-etal-2025-character}.

Yet fully recovering this implicit long-horizon reasoning from novels is difficult, demanding a thorough understanding of a character's traits, their accumulated memory, and the surrounding situation. 
The same history of slights and broken promises may settle into silent resentment in one character yet escalate into a decisive, irreversible act in another, and the memories behind such a choice do not always point the same way: they may reinforce one another, cut against one another, or bear on the decision in different respects. 
Supervised fine-tuning can only imitate a single reference trajectory, and its training signal never lands on the model's own generations, leaving the failure modes specific to its own reasoning unaddressed~\citep{chu2025sft}.
Reinforcement learning offers an alternative, but our setting provides little to reward reliably: open-ended role-play admits no verifiable answer, and an identical surface decision may follow from a weighing of past memories or merely coincide with it. 

The key missing supervision is therefore not whether an agent can access memory, but how it should transform memory into a decision. 
Modeling this transformation requires addressing two challenges at once: 
first, the model must learn a reasoning schema that explains how accumulated, often conflicting memories shape the character's current motivations, constraints, and behavioral tendencies; 
second, training must provide supervision over this intermediate reasoning process rather than only judging the final response.

To address these challenges, we propose \textsc{MemRole}, a two-stage framework that makes the memory-to-decision process explicit and improves it through critique-guided on-policy self-distillation.
First, supervised cold-start training teaches \textbf{L}ong-\textbf{H}orizon \textbf{M}emory \textbf{T}hinking (LHMT), a three-stage schema of context understanding, memory activation, and response strategy planning (Figure~\ref{fig:memory_reasoning}) that makes explicit how past memories shape the current decision rather than serving merely as static context. 
Second, \textbf{S}elf-\textbf{R}efining \textbf{O}n-\textbf{P}olicy \textbf{S}elf-\textbf{D}istillation (SR-OPSD) provides adaptive reasoning-level supervision: the model critiques its own on-policy rollouts with a data-driven rubric, and a critique-conditioned teacher view distills token-level guidance back into the student. 
In this way, \textsc{MemRole} supervises the intermediate memory-to-decision process directly, without reducing feedback to a scalar objective over the final response. The reference enters only through the critique, never as an imitation target.

To validate \textsc{MemRole}, we construct \textsc{MemRole-Dataset}, a decision-centered role-playing dataset built from high-quality novels. 
We segment each novel into scenes and build, for each major character, a memory pool of prior-scene nodes rewritten from their perspective. 
Anchored to the scene's position on the narrative timeline, each pool contains only past scenes the character could plausibly know, separating what they know from what they do not.
Trained on this data, \textsc{MemRole} substantially improves memory-grounded role-playing over strong role-playing baselines while remaining competitive with frontier LLMs.

Our contributions are threefold. 
(1) We propose \textsc{MemRole}, a framework that reasons over long-horizon memories to make role-consistent decisions, combining LHMT with a two-stage training procedure of SFT followed by SR-OPSD. 
(2) We construct \textsc{MemRole-Dataset}, a decision-centered role-playing dataset with character-grounded, timeline-constrained memory pools. 
(3) Controlled ablations show that every LHMT stage and every component of SR-OPSD contributes, with the largest gains concentrated in memory activation and in the data-driven rubric.

\section{Related Work}
\paragraph{LLM-based Role-Playing Agents}
Prior studies mainly relied on prompt engineering or supervised fine-tuning (SFT) to imitate character-specific styles and background knowledge~\citep{shao2023character, wang2025coser}. 
Since surface-level imitation is insufficient for maintaining character consistency in complex scenarios, recent studies have explored reasoning-augmented training for RPAs, including structured reasoning in CogDual and HER~\citep{liu2025cogdual, du2026her} and RL-based optimization in MOA and RAIDEN-R1~\citep{liao2025moa, wang2025raiden, xu2026improving}. 
These methods improve character imitation and response consistency, but largely reason within the current turn rather than over a character's accumulated history.

\paragraph{Memory for Role-Playing Agents}
To sustain consistency across scenes, prior work equips role-playing agents with character memories, typically by retrieving past snippets and injecting them as supplementary context~\citep{wang2024rolellm, ahn-etal-2024-timechara, cheng2026psymem, ji2026ripplemem}. 
Existing methods typically treat memory as a flat collection for context injection, rather than a logical basis for reasoning about the character's state. 
We instead build per-character, timeline-constrained memory pools and train the model to reason explicitly over how these memories shape a character's present decision.

\paragraph{On-Policy Distillation}
Knowledge distillation transfers capabilities from teacher to student models, and recent on-policy variants supervise the student on its own generated trajectories to mitigate the train--test distribution mismatch~\citep{gu2024minillm, agarwal2024policy, song2026survey}. 
A closely related line internalizes reasoning through information-asymmetric self-distillation, where one model acts as both student and teacher and the asymmetry arises from differing input contexts rather than parameters~\citep{zhao2026self, he2026self}. 
We adapt this paradigm to long-horizon memory reasoning: rather than exposing reference solutions directly, we condition the teacher view on per-rollout critiques distilled from the reference, yielding fine-grained supervision for memory-guided role-playing decisions.

\section{\textsc{MemRole} Dataset}
\label{sec:data-curation}

\subsection{Why the \textsc{MemRole} Dataset Is Needed}
\label{sec:why-dataset}

Long-horizon memory reasoning imposes three requirements on the underlying data, none of which is jointly satisfied by existing role-playing corpora (Table~\ref{tab:dataset-compare}).
First, memory must be \emph{cross-scene and traceable}: a character's present decision is driven by events from many prior scenes, so memory cannot be reduced to a static profile~\citep{wang2024rolellm,lu2024large} or to the local surrounding context~\citep{wang2025coser,du2026her}.
Second, scenes must be \emph{decision-centered}: studying how memory becomes decision requires turns where a character makes consequential, character-driven choices, taking a stance, selecting an action, resolving a value conflict, or reaching a turning point, rather than turns dominated by incidental exchange.
Third, memory must respect the character's \emph{epistemic boundary}: a character can only reason from what it could plausibly know at the current narrative point, a constraint absent from prior datasets that retrieve by surface relevance~\citep{wang2025coser} or align responses with a static character graph~\citep{cheng2026psymem}.

As summarized in Table~\ref{tab:dataset-compare}, prior datasets satisfy at most one of these requirements; profile- and attribute-alignment corpora align with static character dimensions rather than reasoning over how past events shape present choices, and decision-oriented benchmarks~\citep{xu-etal-2025-character} evaluate the final choice while supervising neither memory nor reasoning.
Other lines are orthogonal to our goal: large-scale synthetic corpora~\citep{he2025crab, yang2024crafting} prioritize role coverage over traceable narrative memory, while general-domain and multi-agent frameworks~\citep{gao2025tailorrpa,xu2026adamarp} target hallucination control and scene orchestration rather than memory-driven decision-making.

\subsection{Data Curation}
\label{sec:curation}

\begin{figure*}[t]
    \centering
    \includegraphics[width=\textwidth]{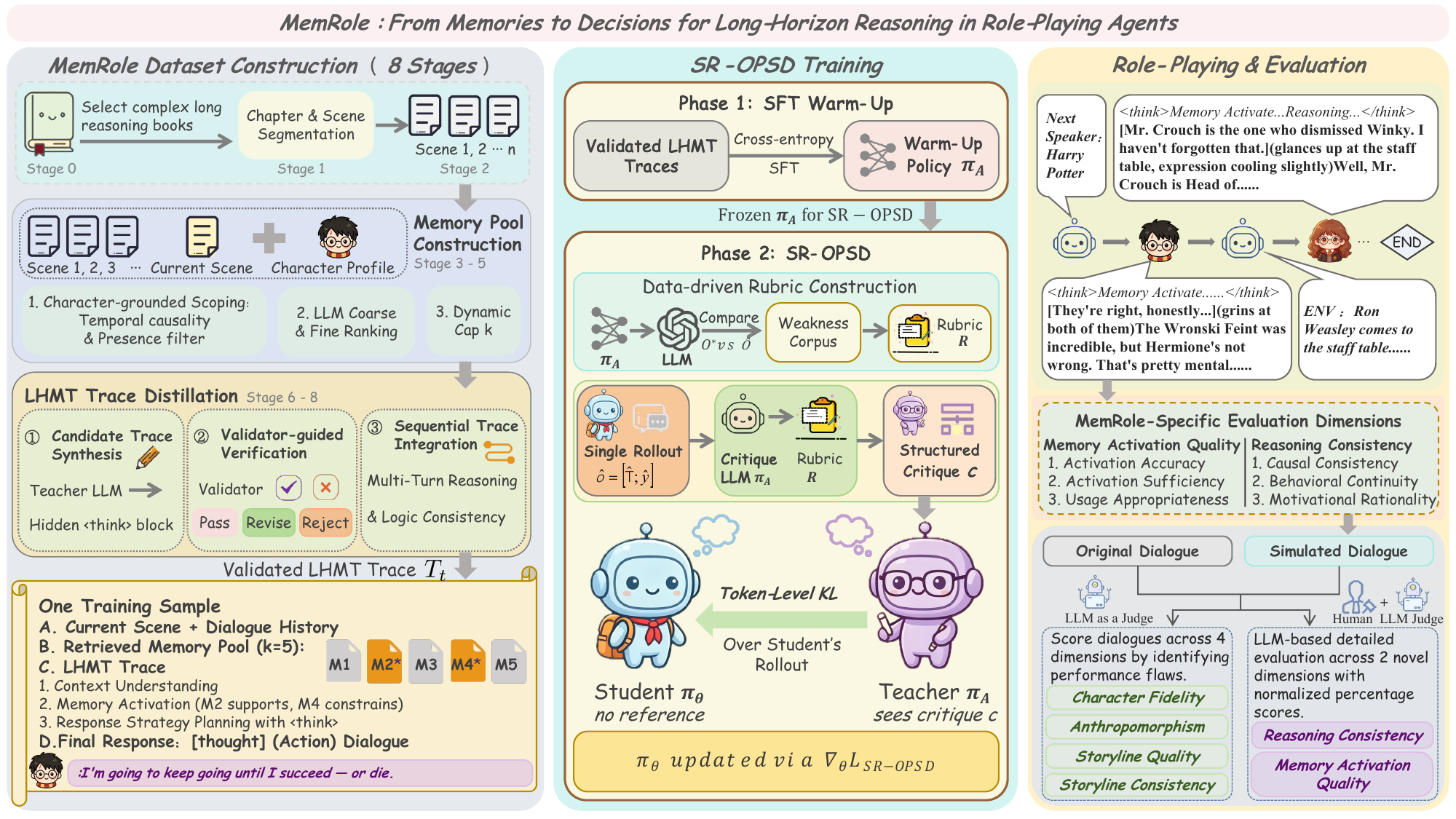}
     \caption{Overview of our framework. \textbf{Left (Data Construction):} The pipeline for constructing the \textsc{MemRole} dataset, involving chapter-scene segmentation, decision-dense scene filtering, and character-grounded memory pool creation. \textbf{Middle (Training):} A two-stage optimization procedure comprising \textbf{SFT warm-up} on validated reasoning traces and \textbf{SR-OPSD refinement} (bottom), which aligns the model with critique-aware teacher distributions. \textbf{Right (Evaluation):} Simulated role-playing and multi-dimensional evaluation on the \textsc{MemRole} Test benchmark.}
    \label{fig:main}
\end{figure*}

\textsc{MemRole} is constructed from high-quality novels through a multi-stage pipeline (Figure~\ref{fig:main}, left) that yields \emph{scene--focal character} instances, each pairing a decision-dense scene with the focal character's profile and a per-character, timeline-anchored memory pool; the training target is the LHMT trace followed by the visible response.
We first segment each novel into chapters and scenes and extract structured scene information, dialogue, and character-centric metadata; all interaction turns are extracted from the original text rather than freely generated, preserving narrative authenticity.
Two design choices make the resulting instances decision-bearing rather than incidental: decision-density filtering and decision-outcome masking.

\paragraph{Decision-density filtering.} To ensure that every instance carries a genuine decision signal, we use an LLM to score whether the focal character makes explicit, character-driven decisions in a scene, and retain only samples above a decision-density threshold. This prioritizes scenes with stance-taking, refusals, commitments, and narrative turning points over filler dialogue.
\paragraph{Decision-outcome masking.} Role-playing setups commonly place a scene introduction and character motivation in the system prompt, which often already disclose the focal character's eventual choice and the scene's outcome. 
A model can then reproduce the behavior from the prompt rather than reason toward it. 
To ensure the response is genuinely \emph{inferred} rather than recoverable from the prompt, we clean the model-visible context so that it retains only the situational conditions and other characters' triggering behaviors, while removing any content that reveals the focal character's decision, its motivation-as-given, or the resulting outcome. 

Additional details on the dataset and its construction pipeline are provided in Appendix~\ref{app:dataset}.

\subsection{Memory Pool Construction}
\label{sec:memory-pool}

The memory pool is the core data structure of \textsc{MemRole}.
For each scene--focal character pair, we construct a dedicated pool of $N$ nodes, where each node corresponds to a prior scene that may shape the focal character's current motivations, beliefs, relationships, or behavioral tendencies.
Each node is an LLM-generated, \emph{character-perspective} summary linked to its source \texttt{scene\_id}, so that every memory is traceable to the original narrative and reflects what that character experienced rather than an omniscient account.

\paragraph{Candidate scoping.} To keep the pool within the character's epistemic boundary, candidate scenes must satisfy two constraints: \emph{temporal validity}---the scene occurs before the current scene, preventing future-knowledge leakage; and \emph{character grounding}---the focal character is physically present, not merely mentioned by others. Together these ensure the pool contains only information the character could plausibly know at the current narrative stage.

\paragraph{Memory selection.} Within the scoped candidates, we select nodes via a two-step LLM-assisted process: a coarse ranking that produces an initial relevant list, followed by a fine ranking based on pairwise comparisons. To let the pool grow with the candidate set while avoiding inflation, we retain the top-$k$ candidates under a dynamic cap:
\begin{equation}
k = \min\!\left(N,\ \big\lceil b + \alpha \cdot \max(0,\, N - b)\big\rceil\right),
\end{equation}
where $b$ is a base size guaranteeing enough memories for small candidate sets and $\alpha \in (0,1)$ is a sub-linear growth rate that lets the pool expand slowly as candidates accumulate. 
A sensitivity analysis of the retrieval budget is provided in Appendix~\ref{app:memory_budget}.

\subsection{Dataset Statistics}
\label{sec:statistics}

\textsc{MemRole} comprises 75 novels, 2,000 character profiles, 14,844 scenes, and 181,137 role-playing dialogue turns (Table~\ref{tab:memrole_overview}).
To prevent narrative leakage, we split at the book level so that all scenes from the same novel fall into the same split.
On average, each turn activates 0.721 memories; among turns with activated memories, 65.03\% integrate multiple memories whose effects are \emph{aligned}, \emph{opposed}, or \emph{mixed} (Table~\ref{tab:memory_effect_pattern_distribution}), highlighting the
need for multi-memory integration beyond single-event retrieval.

\section{Method}
\label{sec:method}

\subsection{Long-Horizon Memory Thinking}
\label{sec:lhmt}

\textsc{MemRole} addresses this need with Long-Horizon Memory Thinking (LHMT), a decision-oriented reasoning schema that makes the memory-to-decision process explicit. 
As shown in Figure~\ref{fig:main}(left) , LHMT decomposes long-horizon memory reasoning into three ordered components: \emph{context understanding}, \emph{memory activation}, and \emph{response strategy planning}, which are written as natural-language paragraphs in an auxiliary \texttt{<think>} block before the visible character response.

\paragraph{Context Understanding.}
This stage analyzes the character's current state, the interactional demands of the dialogue, and the character-profile information relevant to the current turn, thereby clarifying what the response needs to accomplish.

\paragraph{Memory Activation.}
This stage determines whether the current turn needs to consult the memory pool. If so, the model selects memories relevant to the current decision and reasons about how they shape the character's motivations, constraints, and behavioral tendencies. When multiple memories are relevant, the model further integrates their aligned, conflicting, or complementary effects to derive a memory-grounded decision tendency for the current turn.

\paragraph{Response Strategy Planning.}
Given input $x=(C_t, P, M_{r,s})$, where $C_t$ is the dialogue context, $P$ the character profile, and $M_{r,s}=\{m_1,\dots,m_n\}$ the memory pool of character $r$ in scene $s$, the model generates $o_t=[T_t; y_t]\sim\pi_\theta(\cdot\mid x)$, i.e., an LHMT trace $T_t$ followed by the visible response $y_t$.

\subsection{Self-Refining On-Policy Self-Distillation for LHMT}
\label{sec:opsd}

\paragraph{SFT warm-up with synthesized LHMT traces.}
Since LHMT traces are not available in the original novel corpus, we synthesize them with a generator--validator distillation pipeline (Figure~\ref{fig:main}, left). 
For each focal-character turn, a generator LLM produces a candidate LHMT trace $T_t$ grounded in the dialogue context, role prompt, memory pool, and original response $y_t$, without copying the response into the trace. 

A validator LLM filters candidates with a validation checklist covering structural validity, memory grounding, memory-need consistency, and leakage risks.
Validated traces are used for supervised warm-up, with further details in Appendix~\ref{app:trace_construction}.

We first perform supervised fine-tuning on a mixture of LHMT-augmented role-playing trajectories and general instruction-tuning data, as detailed in Appendix~\ref{app:sft_mixture}. For an LHMT-supervised example with input $x=(C_t, P, M_{r,s})$ and target output $o^*=[T_t;y_t]$, the objective is the standard next-token negative log-likelihood:
\begin{equation}
\mathcal{L}_{\mathrm{SFT}}
=
-\sum_{j=1}^{|o^*|}
\log \pi_{\theta}(o^*_j \mid x, o^*_{<j}).
\end{equation}
We denote the resulting model as $\pi_A$. In SR-OPSD, we initialize the trainable policy $\pi_\theta$ from $\pi_A$ and keep another copy of $\pi_A$ frozen as the critique model and teacher view. We empirically validate this choice against alternative critique models in Appendix~\ref{app:critique_model}.

\paragraph{Data-driven Rubric Construction.}
SR-OPSD uses a critique rubric to diagnose weaknesses in the policy's own generations. Rather than manually specifying the rubric, we induce it from the empirical error patterns of the SFT warm-up model $\pi_A$(Figure~\ref{fig:main}, middle). Specifically, we sample $M$ rollouts $\{\hat{o}^{(i)}\}_{i=1}^{M}$ from $\pi_A$ and compare each rollout with its reference output $o^{*(i)}$ to collect a weakness corpus $\{w^{(i)}\}_{i=1}^{M}$.

The weakness corpus is then consolidated through an offline hierarchical summarization process into a global pool of $L$ critique dimensions:
\[
\mathcal{R} = \{d_l\}_{l=1}^{L},
\]
where each $d_l$ denotes a recurring weakness dimension. The rubric remains fixed during SR-OPSD training, while the critique model dynamically selects and instantiates the relevant dimensions for each on-policy rollout. Details are provided in Appendix~\ref{app:rubric_construction}.

\paragraph{Critique-aware Rollout Diagnosis.}
Let $\mathcal{D}=\{(x,o^*)\}$ denote the LHMT-supervised training set used for SR-OPSD. In each training iteration, given $(x, o^*) \sim \mathcal{D}$, the policy $\pi_\theta$ first samples an on-policy trajectory:
\[
\hat{o}=[\hat{T};\hat{y}] \sim \pi_\theta(\cdot \mid x),
\]
where $\hat{T}$ is the generated LHMT trace and $\hat{y}$ is the generated role-playing response. The frozen $\pi_A$ then receives $(x,\hat{o},\mathcal{R},o^*)$ and, guided by $\mathcal{R}$, generates a structured critique $c$ that identifies the rubric dimensions in which $\hat{o}$ deviates from $o^*$. Details are provided in Appendix~\ref{app:sropsd_critique_prompt}.

\begin{table*}[t]
\centering
\small
\setlength{\tabcolsep}{3.5pt}
\renewcommand{\arraystretch}{1}
\setlength{\aboverulesep}{0.25ex}
\setlength{\belowrulesep}{0.25ex}
\begin{tabular*}{\textwidth}{
l @{\extracolsep{\fill}} |
c c c c c c c |
c c |
c
}
\toprule
\multirow{2}{*}{\textbf{Model}}
& \multicolumn{7}{c|}{\textbf{MemRole Test}}
& \multicolumn{2}{c|}{\textbf{Life Choice}}
& \multicolumn{1}{c}{\textbf{Cross-MR}} \\
\cmidrule(lr){2-8}
\cmidrule(lr){9-10}
\cmidrule(lr){11-11}
& \textbf{SC}
& \textbf{AN}
& \textbf{CF}
& \textbf{SQ}
& \textbf{MAQ}
& \textbf{RC}
& \textbf{Avg.}
& \textbf{ACC}
& \textbf{+motivation}
& \textbf{ACC} \\
\midrule
Claude-4.6-Opus
& \textbf{75.26} & 63.11 & \textbf{69.84} & 73.77 & 72.95
& \textbf{80.02} & \textbf{72.49} & \textbf{85.14} & 98.43 & {65.18} \\

Gemini-3-Pro
& 74.33 & 45.98 & 63.57 & \textbf{77.22} & 64.53
& 73.86 & 66.58 & 78.77 & 97.41 & {63.54} \\

DeepSeek-v4-Pro
& 66.37 & 59.76 & 61.14 & 70.96 & 69.78
& 70.33 & 66.39 & 74.58 & 97.32 & {63.77} \\

GPT-5.4
& 64.76 & 55.28 & 57.95 & 70.78 & 65.89
& 72.01 & 64.45 & 73.69 & 97.35 & {67.12} \\

\textbf{MemRole-32B}
& 61.12 & 51.97 & 56.73 & 70.82 & \textbf{72.98}
& 71.87 & 64.25 & 79.11 & 97.72 & \textbf{69.83} \\
\midrule

Doubao-seed-2.0-Pro
& 51.09 & \textbf{63.35} & 43.97 & 76.13 & 48.54
& 66.12 & 58.20 & 81.24 & 98.03 & {61.64} \\

Gemini-2.5-Pro
& 57.77 & 41.98 & 53.72 & 75.76 & 47.95
& 68.13 & 57.55 & 78.02 & 96.69 & {57.96} \\

Grok-4.2
& 48.04 & 47.66 & 44.89 & 70.43 & 70.88
& 49.96 & 55.31 & 78.82 & 97.79 & {59.33} \\

DeepSeek-v3.2
& 51.73 & 44.03 & 43.59 & 62.21 & 58.49
& 62.36 & 53.74 & 72.84 & 93.57 & {52.12} \\

GLM-5
& 42.34 & 50.27 & 37.15 & 74.48 & 52.26
& 62.38 & 53.15 & 81.46 & \textbf{98.47} & {51.05} \\

HER-RL
& 55.13 & 45.17 & 42.64 & 62.83 & 53.93
& 57.34 & 52.84 & 76.78 & 93.39 & {45.27} \\

\textbf{MemRole-8B}
& 55.75 & 44.31 & 39.97 & 66.84 & 52.98
& 56.66 & 52.75 & 75.17 & 92.53 & {53.69} \\

CoSER-70B
& 49.72 & 51.88 & 47.67 & 60.93 & 47.62
& 51.44 & 51.54 & 79.34 & 97.66 & {63.87} \\
\midrule

Qwen3-32B
& 49.36 & 47.91 & 43.62 & 53.11 & 45.93
& 54.38 & 49.05 & 71.68 & 94.74 & {49.61} \\

MiniMax-M2.5
& 45.95 & 48.75 & 31.22 & 48.79 & 41.19
& 56.23 & 45.36 & 73.52 & 94.01 & {53.33} \\

LLaMA-3.1-70B
& 39.68 & 40.80 & 39.97 & 59.82 & 44.45
& 41.04 & 44.29 & 79.87 & 95.97 & {61.27} \\

CoSER-8B
& 40.56 & 42.87 & 34.76 & 55.51 & 39.77
& 41.17 & 42.44 & 68.11 & 90.82 & {44.92} \\

Qwen3-8B
& 41.31 & 40.09 & 37.16 & 51.21 & 42.03
& 42.38 & 42.36 & 69.86 & 88.93 & {41.36} \\

CharacterGLM-6B
& 37.74 & 43.12 & 31.75 & 49.01 & 33.86
& 34.75 & 38.37 & 63.37 & 86.14 & {30.12} \\
\bottomrule
\end{tabular*}

\caption{
Results of different LLMs on \textsc{MemRole} Test, Life Choice, and Cross-MR (0--100, higher is better).
\textsc{MemRole} Test evaluates Storyline Consistency (SC), Anthropomorphism (AN), Character Fidelity (CF), Storyline Quality (SQ), Memory Activation Quality (MAQ), and Reasoning Consistency (RC).
Life Choice evaluates persona-consistent decision-making with given
motivation, while Cross-MR evaluates character-motivation recognition
accuracy.
}
\label{tab:main_results}
\end{table*}

\paragraph{Asymmetric Self-Distillation.}
SR-OPSD performs information-asymmetric self-distillation on the policy model's own rollout $\hat{o}$ (Figure~\ref{fig:main}, middle). 
The student view predicts each next token using only the original input $x$ and the on-policy prefix $\hat{o}_{<j}$:
\[
p_S^{(j)} = \pi_\theta(\cdot \mid x, \hat{o}_{<j}).
\]
In contrast, the teacher view uses the same on-policy prefix but additionally conditions on the structured critique $c$ as privileged context:
\[
p_T^{(j)} = \pi_A(\cdot \mid x, c, \hat{o}_{<j}).
\]
We then minimize the token-level KL divergence on the policy model's own rollout:
\[
\mathcal{L}_{\mathrm{SR\text{-}OPSD}} = \frac{1}{|\hat{o}|} 
\sum_{j=1}^{|\hat{o}|} \mathrm{KL}\left(p_T^{(j)} \,\|\, p_S^{(j)}\right).
\]
At inference time, the model uses only the original input $x$, without access to $o^*$ or the critique model.
The complete SR-OPSD training procedure is provided in Appendix~\ref{app:sropsd_algorithm}.

\section{Experiments}

\subsection{Experimental Setup}
\paragraph{Models.}
We compare \textsc{MemRole} with strong commercial LLMs, open-source general LLMs, and role-playing baselines. 
\textsc{MemRole} is fine-tuned from Qwen3-8B and Qwen3-32B~\citep{yang2025qwen3}. 
We use Gemini-3-Pro-Preview as the environment model for generating interactive contexts and as the next-speaker prediction (NSP) model. 

\paragraph{Metrics.}
Following CoSER~\citep{wang2025coser}, we use GPT-5.5 to evaluate simulated role-playing conversations across four general role-playing dimensions: \textbf{Storyline Consistency} (SC),
\textbf{Anthropomorphism} (AN), \textbf{Character Fidelity} (CF), and \textbf{Storyline Quality} (SQ).
To better assess long-horizon memory reasoning, we further introduce two \textsc{MemRole}-specific metrics: \textbf{Memory Activation Quality} (MAQ) and \textbf{Reasoning Consistency} (RC) (Figure~\ref{fig:main}, right). 
Memory Activation Quality evaluates activation accuracy, activation sufficiency, and usage appropriateness, while Reasoning Consistency evaluates causal consistency, behavioral continuity, and motivational rationality. 
For LifeChoice and Cross-MR, we report accuracy.
Detailed rubrics are provided in Appendix~\ref{app:eval_details}.

\paragraph{Benchmarks.}
We evaluate models on four complementary benchmarks.
\textsc{MemRole} Test evaluates long-horizon memory-grounded role-playing, while CoSER~\citep{wang2025coser} evaluates general role-playing quality in established-character simulations.
For \textsc{MemRole} Test and CoSER, we evaluate 200 conversations containing 20 interaction turns.
Following the evaluation protocol of CoSER~\citep{wang2025coser}, we score only the externally observable responses in these open-ended conversations, excluding reasoning traces.
To provide complementary and more objective evaluations of character decision-making, we further include two multiple-choice benchmarks.
LifeChoice~\citep{xu2024character} evaluates persona-consistent decision-making on 1,000 multiple-choice instances, while CroSS-MR~\citep{yuan2024evaluating} evaluates character-motivation recognition on all 445 instances.
Detailed dataset statistics and evaluation rubrics are provided in Appendices~\ref{app:dataset} and~\ref{app:eval_details}, respectively.

\subsection{Main Results}
\label{sec:main_results}

As shown in Table~\ref{tab:main_results}, \textsc{MemRole} substantially improves memory-grounded role-playing over its backbone models and prior role-playing baselines. \textsc{MemRole}-32B achieves an average score of 64.25 on \textsc{MemRole} Test, outperforming Qwen3-32B by 15.20 points and obtaining the highest MAQ score among all evaluated models. \textsc{MemRole}-8B also improves over Qwen3-8B by 10.39 points and surpasses both CoSER-8B and CoSER-70B, showing that the gains are not merely due to model scale. 

On CoSER, \textsc{MemRole}-32B achieves competitive overall performance and the highest Storyline Quality score in our CoSER evaluation; full results are reported in Appendix~\ref{app:coser_full_results}. On LifeChoice, \textsc{MemRole}-32B remains competitive, indicating that the proposed memory-reasoning training preserves general persona-driven decision-making ability. On CroSS-MR, \textsc{MemRole}-32B achieves the highest accuracy of 69.83, outperforming Qwen3-32B by 20.22 points.

We further assess evaluation robustness through alternative-judge scoring and pairwise comparisons with randomized response order. Both analyses yield broadly consistent comparative conclusions (Appendices~\ref{app:cross_judge_evaluation} and~\ref{app:pairwise_evaluation}).

\subsection{Human Evaluation}
\label{sec:human_evaluation}

We recruit 20 evaluators familiar with narrative role-playing to assess anonymized responses from six representative LLMs on 100 randomly selected \textsc{MemRole} test samples using a 1--10 scale. As shown in Table~\ref{tab:human_response_evaluation}, \textsc{MemRole}-32B achieves the highest human score (7.02) and pairwise win rate (72.3\%), outperforming Qwen3-32B, HER-RL, and CoSER-70B. \textsc{MemRole}-8B also outperforms Qwen3-8B. We further assess human--automatic alignment on 300 responses and 300 LHMT traces using case-level agreement, Cohen's $\kappa$, and Spearman's $\rho$. The results show strong agreement for both responses (89.4\%, $\kappa=0.77$, $\rho=0.81$) and LHMT traces (90.2\%, $\kappa=0.80$, $\rho=0.83$). Detailed results are provided in Appendix~\ref{app:human_evaluation}

\subsection{Ablation Study}

\paragraph{Effect of LHMT components.}
We ablate each LHMT component to assess its contribution. As shown in Table~\ref{tab:ablation}, removing any component degrades performance. Removing \textit{memory activation} causes the largest Memory drop ($-10.1$), highlighting its central role in long-horizon grounding. Removing \textit{response strategy planning} mainly hurts Dialogue ($-5.2$), while removing \textit{context understanding} leads to smaller but consistent drops ($-3.2/-3.9$), suggesting that it provides useful scaffolding for later reasoning.

\paragraph{Effect of SR-OPSD training.}
The training-stage ablations show that each part of SR-OPSD contributes to performance. Standard OPSD, which directly conditions the teacher on the full reference, provides modest gains over SFT ($+1.5/+1.4$). Replacing direct reference conditioning with rollout-specific critiques further improves performance ($+1.1/+2.2$), and the data-driven rubric yields the final gain ($+2.3/+4.0$), indicating that rubric-guided critiques provide more targeted supervision for memory-to-decision reasoning.

\begin{table}[t]
\centering
\small

\setlength{\tabcolsep}{3.2pt}
\renewcommand{\arraystretch}{0.86}
\setlength{\aboverulesep}{0.22ex}
\setlength{\belowrulesep}{0.22ex}

\begin{adjustbox}{max width=\columnwidth}
\begin{tabular}{@{}lcc@{}}
\toprule
\textbf{Setting}
& \textbf{Dialogue}
& \textbf{Memory} \\
\midrule

\textbf{MemRole-8B (Full)}
& \textbf{51.7}
& \textbf{54.8} \\

\midrule
\multicolumn{3}{@{}l}{\textit{Effect of LHMT components}} \\
\hspace{0.8em} w/o Context Understanding
& 48.5 & 50.9 \\
\hspace{0.8em} w/o Memory Activation
& 48.6 & 44.7 \\
\hspace{0.8em} w/o Response Strategy Planning
& 46.5 & 52.3 \\

\midrule
\multicolumn{3}{@{}l}{\textit{Effect of training stages}} \\
\hspace{0.8em} SFT only ($\pi_A$)
& 46.8 & 47.2 \\
\hspace{0.8em} Standard OPSD~\citep{zhao2026self}
& 48.3 & 48.6 \\
\hspace{0.8em} SR-OPSD w/o rubric
& 49.4 & 50.8 \\

\bottomrule
\end{tabular}
\end{adjustbox}

\caption{%
Ablation study on the \textsc{MemRole} test split using the 8B backbone.
Dialogue averages SC, AN, CF, and SQ; Memory averages MAQ and RC.
Higher is better.
}
\label{tab:ablation}
\end{table}

\subsection{Analysis of LHMT Reasoning}
\label{sec:lhmt_reasoning_analysis}

We analyze LHMT in terms of schema effectiveness, supervision-trace
validity, and the quality of model-generated reasoning traces.

\paragraph{Effectiveness of the LHMT Schema.}
We evaluate LHMT at both prompting and training levels. For prompting, we conduct a zero-shot experiment on frozen Qwen3 models, comparing Direct Response (no explicit reasoning), Generic CoT (free-form step-by-step reasoning), and LHMT Prompting (three-stage structured reasoning). As shown in Table~\ref{tab:lhmt_effectiveness}(a), LHMT performs comparably to Generic CoT on 8B but shows a greater advantage on 32B. For training, we compare Direct-response SFT, Generic-CoT SFT, and LHMT SFT under matched settings, using identical instances, target responses, data mixtures, and optimization configurations. Generic-CoT traces use the same teacher and generation setup as LHMT traces, differing only in reasoning structure. Table~\ref{tab:lhmt_effectiveness}(b) shows that Generic-CoT SFT improves over Direct-response SFT by 1.56, 1.67, and 2.33 points in Dialogue Avg., MAQ, and RC, respectively. LHMT SFT yields further gains of 1.06, 1.71, and 1.49 points, respectively. Additional results across Dense RAG and GraphRAG are provided in Appendix~\ref{app:memory_backend_generalization}.

\begin{table}[!t]
\centering
\small

\setlength{\tabcolsep}{3.5pt}
\renewcommand{\arraystretch}{0.88}
\setlength{\aboverulesep}{0.25ex}
\setlength{\belowrulesep}{0.25ex}

\begin{tabular}{@{}llccc@{}}
\toprule
\multicolumn{5}{c}{%
\textbf{(a) Prompting-level comparison}
} \\
\midrule
\textbf{Model}
& \textbf{Reasoning Mode}
& \makecell{\textbf{Dialogue}\\[-1pt]\textbf{Avg.}}
& \textbf{MAQ}
& \textbf{RC} \\
\midrule

\multirow{3}{*}{Qwen3-8B}
& Direct Response
& 42.44 & 42.03 & 42.38 \\
& Generic CoT
& \textbf{43.93} & 44.26 & 43.11 \\
& LHMT Prompting
& 43.87 & \textbf{44.78} & \textbf{44.34} \\

\midrule

\multirow{3}{*}{Qwen3-32B}
& Direct Response
& 48.50 & 45.93 & 54.38 \\
& Generic CoT
& 49.85 & 48.46 & 58.81 \\
& LHMT Prompting
& \textbf{52.03} & \textbf{51.58} & \textbf{59.05} \\

\midrule
\multicolumn{5}{c}{%
\textbf{(b) Training-level comparison}
} \\
\midrule
\textbf{Backbone}
& \textbf{Supervision Format}
& \makecell{\textbf{Dialogue}\\[-1pt]\textbf{Avg.}}
& \textbf{MAQ}
& \textbf{RC} \\
\midrule

\multirow{3}{*}{Qwen3-8B}
& Direct-response SFT
& 44.17 & 42.84 & 44.36 \\
& Generic-CoT SFT
& 45.73 & 44.51 & 46.69 \\
& \textbf{LHMT SFT}
& \textbf{46.79}
& \textbf{46.22}
& \textbf{48.18} \\

\bottomrule
\end{tabular}

\caption{%
Prompting- and training-level effectiveness of LHMT.
Dialogue Avg.\ denotes the average of SC, AN, CF, and SQ.
Higher is better.
}
\label{tab:lhmt_effectiveness}
\end{table}

\paragraph{Quality of Supervision Traces.}
To assess whether LHMT traces reflect narrative-supported decisions rather than plausible post-hoc explanations, we sample 2,000 instances from the training set containing explicit decision evidence, such as internal monologues or stated motivations. Given the current context, candidate memories, their source passages, and the LHMT trace, an automatic judge evaluates Context Faithfulness, Memory Grounding, and Plan Coherence, with acceptance rates of 90.4\%, 93.4\%, and 88.9\%, respectively. A trace is considered fully valid only when all three dimensions meet the quality criteria without contradicting the character's explicit thoughts, knowledge, or narrative timeline. Overall, 84.4\% of the traces meet this standard. Human validation of the judge is reported in Appendix~\ref{app:human_evaluation}.

\paragraph{Quality of Generated Traces.}
\label{sec:lhmt_quality}

We evaluate LHMT trace quality along three dimensions: \textbf{Context Faithfulness}, \textbf{Memory Grounding}, and \textbf{Plan Coherence}, using the same 300 test turns for all models. All three dimensions improve from Base + LHMT prompt through SFT warm-up to SR-OPSD. Memory Grounding remains the weakest after SFT but improves the most under SR-OPSD, becoming comparable to the other two. This suggests that critique-aware refinement particularly strengthens memory-to-decision reasoning, consistent with the largest degradation observed when ablating memory activation (Appendix~\ref{app:generated_trace_quality}).

\section{Conclusion}
We presented MemRole, which trains role-playing agents to reason from long-horizon memories to role-consistent decisions by combining the LHMT schema with critique-guided on-policy self-distillation (SR-OPSD), together with a dataset of character-grounded, timeline-constrained memory pools. Experiments and ablations show consistent gains over backbones and role-playing baselines, highlighting the value of supervising intermediate memory reasoning.

\section*{Limitations}
\textsc{MemRole} has two main limitations.
First, LHMT trace synthesis, critique-rubric construction, and automatic evaluation rely on LLM assistance. 
We mitigate potential noise with structured prompts, independent validation, and human expert calibration, while stronger validators or broader human auditing could further improve robustness.
Second, \textsc{MemRole} studies reasoning over pre-existing narrative memories from novels, which provides a controlled setting for memory-to-decision modeling. 
Extending the framework to online memory acquisition and revision during user interaction remains future work.

\section*{Ethics Statement}
This work studies long-horizon memory reasoning for fictional role-playing agents. 
The dataset is constructed from literary narratives and centers on fictional characters, scenes, and character memories; it does not use private user conversations, personally identifiable information, or real-person profiling. 
Because the source novels may contain copyrighted text and socially biased or sensitive portrayals, we will not redistribute raw copyrighted content. 
Instead, we will release only copyright-compliant derived resources, such as annotations, prompts, statistics, and model outputs, for research purposes. 
We also acknowledge that stronger role-playing agents may be misused for deceptive impersonation, emotionally manipulative interaction, or unsafe fictional simulation. 
Therefore, systems based on this work should be deployed with clear AI disclosure, content safeguards, and appropriate human oversight. 
This work is intended for research on fictional narrative reasoning, not for simulating real individuals without consent or making decisions that affect real people.

\bibliography{custom}

@article{song2026survey,
  title={A Survey of On-Policy Distillation for Large Language Models},
  author={Song, Mingyang and Zheng, Mao},
  journal={arXiv preprint arXiv:2604.00626},
  year={2026}
}

@inproceedings{gu2024minillm,
  title={Minillm: Knowledge distillation of large language models},
  author={Gu, Yuxian and Dong, Li and Wei, Furu and Huang, Minlie},
  booktitle={The twelfth international conference on learning representations},
  year={2024}
}

@inproceedings{shao2023character,
  title={Character-llm: A trainable agent for role-playing},
  author={Shao, Yunfan and Li, Linyang and Dai, Junqi and Qiu, Xipeng},
  booktitle={Proceedings of the 2023 Conference on Empirical Methods in Natural Language Processing},
  pages={13153--13187},
  year={2023}
}

@inproceedings{wang2025coser,
  title={Coser: Coordinating llm-based persona simulation of established roles},
  author={Wang, Xintao and Wang, Heng and Zhang, Yifei and Yuan, Xinfeng and Xu, Rui and Huang, Jen-tse and Yuan, Siyu and Guo, Haoran and Chen, Jiangjie and Zhou, Shuchang and others},
  booktitle={Forty-second International Conference on Machine Learning},
  year={2025}
}

@inproceedings{liu2025cogdual,
  title={CogDual: Enhancing Dual Cognition of LLMs via Reinforcement Learning with Implicit Rule-Based Rewards},
  author={Liu, Cheng and Lu, Yifei and Ye, Fanghua and Li, Jian and Chen, Xingyu and Ren, Feiliang and Tu, Zhaopeng and Li, Xiaolong},
  booktitle={Proceedings of the 2025 Conference on Empirical Methods in Natural Language Processing},
  pages={27295--27324},
  year={2025}
}

@article{liao2025moa,
  title={MOA: Multi-Objective Alignment for Role-Playing Agents},
  author={Liao, Chonghua and Wang, Ke and Wu, Yuchuan and Huang, Fei and Li, Yongbin},
  journal={arXiv preprint arXiv:2512.09756},
  year={2025}
}

@inproceedings{wang2024rolellm,
  title={Rolellm: Benchmarking, eliciting, and enhancing role-playing abilities of large language models},
  author={Wang, Noah and Peng, Zy and Que, Haoran and Liu, Jiaheng and Zhou, Wangchunshu and Wu, Yuhan and Guo, Hongcheng and Gan, Ruitong and Ni, Zehao and Yang, Jian and others},
  booktitle={Findings of the Association for Computational Linguistics: ACL 2024},
  pages={14743--14777},
  year={2024}
}

@article{wang2025raiden,
  title={Raiden-r1: Improving role-awareness of llms via grpo with verifiable reward},
  author={Wang, Zongsheng and Sun, Kaili and Wu, Bowen and Yu, Qun and Li, Ying and Wang, Baoxun},
  journal={arXiv preprint arXiv:2505.10218},
  year={2025}
}

@article{du2026her,
  title={HER: Human-like Reasoning and Reinforcement Learning for LLM Role-playing},
  author={Du, Chengyu and Wang, Xintao and Chen, Aili and Li, Weiyuan and Xu, Rui and Liu, Junteng and Huang, Zishan and Tian, Rong and Sun, Zijun and Li, Yuhao and others},
  journal={arXiv preprint arXiv:2601.21459},
  year={2026}
}

@article{cheng2026psymem,
  title={Psymem: Fine-grained psychological alignment and explicit memory control for advanced role-playing llms},
  author={Cheng, Xilong and Qin, Yunxiao and Tan, Yuting and Li, Zhengnan and Wang, Ye and Xiao, Hongjiang and Zhang, Yuan},
  journal={Transactions of the Association for Computational Linguistics},
  volume={14},
  pages={510--529},
  year={2026},
  publisher={MIT Press}
}

@article{xu2024character,
  title={Character is destiny: Can role-playing language agents make personadriven decisions},
  author={Xu, Rui and Wang, Xintao and Chen, Jiangjie and Yuan, Siyu and Yuan, Xinfeng and Liang, Jiaqing and Chen, Zulong and Dong, Xiaoqing and Xiao, Yanghua},
  journal={Preprint},
  year={2024}
}

@inproceedings{agarwal2024policy,
  title={On-policy distillation of language models: Learning from self-generated mistakes},
  author={Agarwal, Rishabh and Vieillard, Nino and Zhou, Yongchao and Stanczyk, Piotr and Ramos Garea, Sabela and Geist, Matthieu and Bachem, Olivier},
  booktitle={International Conference on Learning Representations},
  volume={2024},
  pages={21246--21263},
  year={2024}
}

@inproceedings{zhou2024characterglm,
  title={CharacterGLM: Customizing social characters with large language models},
  author={Zhou, Jinfeng and Chen, Zhuang and Wan, Dazhen and Wen, Bosi and Song, Yi and Yu, Jifan and Huang, Yongkang and Ke, Pei and Bi, Guanqun and Peng, Libiao and others},
  booktitle={Proceedings of the 2024 conference on empirical methods in natural language processing: Industry track},
  pages={1457--1476},
  year={2024}
}

@article{zhao2026self,
  title={Self-Distilled Reasoner: On-Policy Self-Distillation for Large Language Models},
  author={Zhao, Siyan and Xie, Zhihui and Liu, Mengchen and Huang, Jing and Pang, Guan and Chen, Feiyu and Grover, Aditya},
  journal={arXiv preprint arXiv:2601.18734},
  year={2026}
}

@article{he2026self,
  title={Self-distillation zero: Self-revision turns binary rewards into dense supervision},
  author={He, Yinghui and Kaur, Simran and Bhaskar, Adithya and Yang, Yongjin and Liu, Jiarui and Ri, Narutatsu and Fowl, Liam and Panigrahi, Abhishek and Chen, Danqi and Arora, Sanjeev},
  journal={arXiv preprint arXiv:2604.12002},
  year={2026}
}

@inproceedings{ahn-etal-2024-timechara,
    title = "{T}ime{C}hara: Evaluating Point-in-Time Character Hallucination of Role-Playing Large Language Models",
    author = "Ahn, Jaewoo  and
      Lee, Taehyun  and
      Lim, Junyoung  and
      Kim, Jin-Hwa  and
      Yun, Sangdoo  and
      Lee, Hwaran  and
      Kim, Gunhee",
    editor = "Ku, Lun-Wei  and
      Martins, Andre  and
      Srikumar, Vivek",
    booktitle = "Findings of the Association for Computational Linguistics: ACL 2024",
    month = aug,
    year = "2024",
    address = "Bangkok, Thailand",
    publisher = "Association for Computational Linguistics",
    url = "https://aclanthology.org/2024.findings-acl.197/",
    doi = "10.18653/v1/2024.findings-acl.197",
    pages = "3291--3325"
}

@inproceedings{xu-etal-2025-character,
    title = "Character is Destiny: Can Persona-assigned Language Models Make Personal Choices?",
    author = "Xu, Rui  and
      Wang, Xintao  and
      Chen, Jiangjie  and
      Yuan, Siyu  and
      Yuan, Xinfeng  and
      Liang, Jiaqing  and
      Chen, Zulong  and
      Xiaoqingdong  and
      Xiao, Yanghua",
    editor = "Christodoulopoulos, Christos  and
      Chakraborty, Tanmoy  and
      Rose, Carolyn  and
      Peng, Violet",
    booktitle = "Findings of the Association for Computational Linguistics: EMNLP 2025",
    month = nov,
    year = "2025",
    address = "Suzhou, China",
    publisher = "Association for Computational Linguistics",
    url = "https://aclanthology.org/2025.findings-emnlp.813/",
    doi = "10.18653/v1/2025.findings-emnlp.813",
    pages = "15038--15059",
    ISBN = "979-8-89176-335-7"
}

@article{conway2000construction,
  title={The construction of autobiographical memories in the self-memory system.},
  author={Conway, Martin A and Pleydell-Pearce, Christopher W},
  journal={Psychological review},
  volume={107},
  number={2},
  pages={261},
  year={2000},
  publisher={American Psychological Association}
}

@incollection{mcadams2011narrative,
  title={Narrative identity},
  author={McAdams, Dan P},
  booktitle={Handbook of identity theory and research},
  pages={99--115},
  year={2011},
  publisher={Springer}
}

@article{chu2025sft,
  title={Sft memorizes, rl generalizes: A comparative study of foundation model post-training},
  author={Chu, Tianzhe and Zhai, Yuexiang and Yang, Jihan and Tong, Shengbang and Xie, Saining and Schuurmans, Dale and Le, Quoc V and Levine, Sergey and Ma, Yi},
  journal={arXiv preprint arXiv:2501.17161},
  year={2025}
}

@article{yang2025qwen3,
  title={Qwen3 technical report},
  author={Yang, An and Li, Anfeng and Yang, Baosong and Zhang, Beichen and Hui, Binyuan and Zheng, Bo and Yu, Bowen and Gao, Chang and Huang, Chengen and Lv, Chenxu and others},
  journal={arXiv preprint arXiv:2505.09388},
  year={2025}
}

@inproceedings{lu2024large,
  title={Large language models are superpositions of all characters: Attaining arbitrary role-play via self-alignment},
  author={Lu, Keming and Yu, Bowen and Zhou, Chang and Zhou, Jingren},
  booktitle={Proceedings of the 62nd Annual Meeting of the Association for Computational Linguistics (Volume 1: Long Papers)},
  pages={7828--7840},
  year={2024}
}

@inproceedings{he2025crab,
  title={Crab: A novel configurable role-playing llm with assessing benchmark},
  author={He, Kai and Huang, Yucheng and Wang, Wenqing and Ran, Delong and Sheng, Dongming and Huang, Junxuan and Lin, Qika and Xu, Jiaxing and Liu, Wenqiang and Feng, Mengling},
  booktitle={Proceedings of the 63rd Annual Meeting of the Association for Computational Linguistics (Volume 1: Long Papers)},
  pages={15030--15052},
  year={2025}
}

@article{yang2024crafting,
  title={Crafting customisable characters with LLMs: A persona-driven role-playing agent framework},
  author={Yang, Bohao and Liu, Dong and Xiao, Chenghao and Zhao, Kun and Tang, Chen and Li, Chao and Yuan, Lin and Yang, Guang and Lin, Chenghua},
  journal={arXiv preprint arXiv:2406.17962},
  year={2024}
}

@article{xu2026adamarp,
  title={Adamarp: An adaptive multi-agent interaction framework for general immersive role-playing},
  author={Xu, Zhenhua and Chen, Dongsheng and Wang, Shuo and Li, Jian and Wang, Chengjie and Han, Meng and Wang, Yabiao},
  journal={arXiv preprint arXiv:2601.11007},
  year={2026}
}

@inproceedings{gao2025tailorrpa,
  title={TailorRPA: A Retrieval-Based Framework for Eliciting Personalized and Coherent Role-Playing Agents in General Domain},
  author={Gao, Zhenpeng and Xin, Xiaofen and Xu, Xiangmin},
  booktitle={Findings of the Association for Computational Linguistics: EMNLP 2025},
  pages={5381--5412},
  year={2025}
}

@inproceedings{yuan2024evaluating,
  title={Evaluating character understanding of large language models via character profiling from fictional works},
  author={Yuan, Xinfeng and Yuan, Siyu and Cui, Yuhan and Lin, Tianhe and Wang, Xintao and Xu, Rui and Chen, Jiangjie and Yang, Deqing},
  booktitle={Proceedings of the 2024 Conference on Empirical Methods in Natural Language Processing},
  pages={8015--8036},
  year={2024}
}

@article{pillemer2003directive,
  title={Directive functions of autobiographical memory: The guiding power of the specific episode},
  author={Pillemer, David},
  journal={Memory},
  volume={11},
  number={2},
  pages={193--202},
  year={2003},
  publisher={Taylor \& Francis}
}

@article{graesser1994constructing,
  title={Constructing inferences during narrative text comprehension.},
  author={Graesser, Arthur C and Singer, Murray and Trabasso, Tom},
  journal={Psychological review},
  volume={101},
  number={3},
  pages={371},
  year={1994},
  publisher={American Psychological Association}
}

@article{zhang2026psymembehav,
  title={PsyMemBehav: Psychology-Informed and Memory-Grounded Two-Level Identity-to-Action Reasoning for Social-Media Behavior Simulation},
  author={Zhang, Wenji and Cheng, Xilong and Qin, Yunxiao and Tu, Mai and Zhu, Tongye and Zhou, Yuhao and Ji, JingBo and Tan, Yuting and Li, Duoyin and Xiao, Hongjiang and others},
  journal={arXiv preprint arXiv:2610.14570},
  year={2026}
}

@article{ji2026ripplemem,
  title={RippleMem: From Isolated Retrieval to Associative Recollection for Long-Term Agent Memory},
  author={Ji, Jingbo and Li, Lingyi and Cheng, Xilong and Zhou, Yuhao and Zhang, Wenji and Tan, Yuting and Qin, Yunxiao},
  journal={arXiv preprint arXiv:2608.13334},
  year={2026}
}

@inproceedings{10.1145/3770855.3818027,
author = {Xiao, Zhihao and Li, Mengting and Wang, Xintao and Li, Linfeng and Shui, Limin and Ji, Mengqi and Cai, Borui},
title = {DREAM: LLM-based Dynamic Role-playing via Event-Aware Memory Graph},
year = {2026},
isbn = {9798400722592},
publisher = {Association for Computing Machinery},
address = {New York, NY, USA},
url = {https://doi.org/10.1145/3770855.3818027},
doi = {10.1145/3770855.3818027},
booktitle = {Proceedings of the 32nd ACM SIGKDD Conference on Knowledge Discovery and Data Mining V.2},
pages = {5686–5697},
numpages = {12},
location = {Jeju Island, Republic of Korea},
series = {KDD '26}
}

@article{xu2026improving,
  title={Improving General Role-Playing Agents via Psychology-Grounded Reasoning and Role-Aware Policy Optimization},
  author={Xu, Zhenhua and Chen, Dongsheng and Li, Jian and Lin, Yitong and Wang, Zhebo and Wu, Jiafu and Jin, Yizhang and Wang, Chengjie and Han, Meng and Wang, Yabiao},
  journal={arXiv preprint arXiv:2606.27025},
  year={2026}
}

\clearpage
\appendix

\section{Dataset}
\label{app:dataset}

\subsection{Comparison with Existing Datasets}
\label{app:dataset_comparison}

Table~\ref{tab:dataset-compare} compares \textsc{MemRole-Dataset} with representative role-playing datasets in terms of memory representation, knowledge-boundary constraints, decision focus, and reasoning supervision.

\begin{table*}[t]
\centering
\resizebox{\textwidth}{!}{
\begin{tabular}{lccccc}
\toprule
Dataset & Source & Memory Representation & \makecell{Knowledge-boundary\\Constraint} & \makecell{Decision\\Focus} & Reasoning Supervision \\
\midrule
RoleLLM~\citep{wang2024rolellm} & Scripts / GPT-gen. & Profile + synthetic QA & -- & \xmark & None \\
DITTO~\citep{lu2024large} & Wikidata (self-aligned) & Model-internal knowledge & -- & \xmark & None \\
CoSER~\citep{wang2025coser} & Novels & Conversations / plots / exp. & \xmark & \xmark & None \\
HER~\citep{du2026her} & Novels (CoSER-derived) & Dialogue history + scenario & \xmark & \xmark & Dual-layer thinking \\
PsyMem~\citep{cheng2026psymem} & Novels & Character / event graph & \xmark & \xmark & None \\
LifeChoice~\citep{xu-etal-2025-character} & Novels & Preceding context & Partial & Choice-level & None \\
\midrule
\textsc{MemRole-Dataset} & Novels & \makecell{Per-character, perspective-\\rewritten, timeline-anchored pool} & \cmark & \cmark & \makecell{LHMT memory-to-\\decision trace} \\
\bottomrule
\end{tabular}
}
\caption{Comparison between \textsc{MemRole} and representative role-playing datasets.
\textbf{Knowledge-boundary Constraint} denotes whether memory is restricted to what the focal
character could plausibly know at the current narrative point, via \emph{temporal validity}
(only prior scenes) and \emph{character grounding} (the focal character must be present, not
merely mentioned). CoSER additionally supports optional retrieval over its book-derived data;
LifeChoice provides partial preceding context but no character-grounded memory pool.
``--'' means the dataset has no cross-scene memory for the constraint to apply.}
\label{tab:dataset-compare}
\end{table*}

\subsection{Dataset Overview}

\textsc{MemRole} is a long-horizon memory-based role-playing dataset
constructed from raw novel corpora. As shown in
Table~\ref{tab:memrole_overview}, the final dataset contains 75 novels,
2,000 character profiles, 14,844 scenes, and 181,137 role-playing
dialogue turns. Each scene contains an average of 12.20 dialogue turns,
and each turn activates 0.721 memories on average. Each sample is
centered on a focal character in a specific scene and includes the
current dialogue context, the original character response, the
character profile, and a long-horizon memory pool.

We split the dataset at the book level to prevent narrative leakage
between training and test sets. The final split contains 64 training
novels and 11 test novels, with all scenes, characters, and dialogue
samples from the same novel assigned to the same split.

We further report memory-use statistics.
Table~\ref{tab:memrole_memory_distribution} shows the distribution of
the number of activated memories per role-playing turn.

Among turns with activated memories, we additionally analyze their
effect patterns. Table~\ref{tab:memory_effect_pattern_distribution}
reports the corresponding distribution.

\begin{table*}[t]
\centering
\small
\setlength{\tabcolsep}{6pt}
\begin{tabular}{lrrrrrr}
\toprule
Split & Novels & Characters & Scenes & Turns & Turns/Scene & Memories/Turn \\
\midrule
Train & 64 & 1,755 & 13,172 & 161,027 & 12.22 & 0.722 \\
Test & 11 & 245 & 1,672 & 20,110 & 12.03 & 0.710 \\
\midrule
Total & 75 & 2,000 & 14,844 & 181,137 & 12.20 & 0.721 \\
\bottomrule
\end{tabular}
\caption{Overview and book-level train/test split statistics of \textsc{MemRole}.}
\label{tab:memrole_overview}
\end{table*}

\begin{table}[t]
\centering
\small
\begin{tabular}{lrr}
\toprule
Activated Memories & Count & Percentage \\
\midrule
0 & 113,687 & 62.8\% \\
1 & 23,590 & 13.0\% \\
2 & 28,432 & 15.7\% \\
3+ & 15,428 & 8.5\% \\
\midrule
Total & 181,137 & 100.0\% \\
\bottomrule
\end{tabular}
\caption{Distribution of activated memories per role-playing dialogue turn in \textsc{MemRole}.}
\label{tab:memrole_memory_distribution}
\end{table}

\begin{table}[t]
\centering
\small
\setlength{\tabcolsep}{6pt}
\begin{tabular}{lrr}
\toprule
Memory Effect Pattern & Count & Percentage \\
\midrule
\textit{single}  & 23,590 & 34.97\% \\
\textit{aligned} & 35,424 & 52.52\% \\
\textit{opposed} & 3,683    & 5.46\% \\
\textit{mixed}   & 4,753    & 7.05\% \\
\midrule
Total & 67,450 & 100.00\% \\
\bottomrule
\end{tabular}
\caption{Distribution of memory-effect patterns among turns with activated memories in \textsc{MemRole}.}
\label{tab:memory_effect_pattern_distribution}
\end{table}

\subsection{Dataset Construction Pipeline}
We curate \textsc{MemRole} through a multi-stage extractive pipeline that goes from raw novel text to a structured, queryable representation. The details are as follows.

\paragraph{Source Selection.}
To ensure narrative quality and character depth, we start from the top 1,000 books in Goodreads' ``Best Books Ever'' list and use an LLM-based screening process to select novels better suited for long-horizon role-playing decision modeling. We prioritize books with sustained character arcs, decision-rich interactions, and cross-scene causal chains.

After this screening process, we obtain 75 novels for dataset construction. The selected novels cover a broad range of genres, including fantasy, historical fiction, science fiction, romance, mystery, literary fiction, adventure, young adult fiction, dystopian fiction, and gothic fiction. This genre coverage introduces diverse narrative styles, social settings, interpersonal conflicts, and character decision patterns, making \textsc{MemRole} suitable for studying long-horizon role-playing behavior across varied story worlds.

\paragraph{Chapter and Scene Segmentation.}
We first segment each book into chapter-level chunks based on its chapter structure. Chapter boundaries are jointly identified by rules and LLMs, and then globally adjudicated to produce a unified chapter list with chapter-level summaries. Within each chapter, we use LLMs to extract candidate scenes and their metadata, followed by global adjudication to obtain non-overlapping scene segments that fully cover the chapter.

\paragraph{Data Extraction.}
We use LLMs to extract structured information from each scene, including location, in-story time, atmosphere, relevant characters, preceding context, scene summary, scene outcome, and dialogue content. Following the message representation in CoSER, we further organize dialogue content into three types of information: character speech, actions, and thoughts.

To filter decision-dense samples, we compute heuristic indicators such as dialogue density, action/thought ratios, and the number of interaction turns, and assign each scene an interaction-type label accordingly. We then combine these labels with the LLM-assessed role-playing value to select high-quality scenes with sufficient character interaction, explicit behavioral choices, and strong plot-advancing effects.

\paragraph{Character Data Organization.}
Based on the extracted scenes and dialogues, we construct a character profile for each major character. We first perform alias resolution with LLMs, mapping different character mentions to a canonical name. Each character profile is then organized into three core components: a \textbf{static profile}, including personality traits, core values, speech style, and behavioral patterns; \textbf{relationships}, recording relation types, relationship trajectories, and the \texttt{scene\_id}s of several key scenes; and a \textbf{dynamic character arc}, which partitions the character's narrative trajectory into 3--7 stages.

Each stage is associated with a \texttt{scene\_id} range, a set of key turning-point \texttt{scene\_id}s, and a state snapshot describing the character's situation, goals, known plot facts, and important relationship states during that stage. This stage-wise structure makes character memory queryable: given any decision point, we can trace back to the preceding character stages and further locate relevant scenes for memory retrieval.

\subsection{Example Instance}

Table~\ref{tab:memrole_example} shows a representative test-set instance with dense memory activation. This example is selected from \textit{Stones from the River}, where the current role-playing turn activates nine past-scene memories.

\begin{table*}[!t]
\centering
\small
\setlength{\tabcolsep}{6pt}
\renewcommand{\arraystretch}{1.15}
\begin{tabular}{>{\bfseries\raggedright\arraybackslash}p{0.17\textwidth}p{0.76\textwidth}}
\toprule
\multicolumn{2}{c}{\textbf{\textit{Stones from the River}} \hfill \textbf{(From Test Set)}} \\
\midrule
Character &
Trudi Montag \\
\midrule
Topic &
Comforting Konrad before his wartime relocation. \\
\midrule
Plot Summary &
Leo Montag prepares a rare formal dinner for Trudi, Eva, Frau Neimann, and Konrad during wartime scarcity. During the meal, he reveals that Konrad and his mother must leave that night for a safer hiding place. Trudi has known this news in advance but has hidden it from Konrad to avoid frightening him. As the departure approaches, she uses the invented story of the Island of the Little People to comfort him while privately struggling with grief, anger, and attachment. \\
\midrule
Scenario &
The scene takes place in Leo Montag's home, including the kitchen and attached pay-library. The atmosphere is bittersweet and tense: the dinner is warm and carefully prepared, but it is also a farewell. Konrad asks why the imagined island in Trudi's story has changed and why its beautiful world has collapsed. The question forces Trudi to respond without exposing the real danger behind his relocation or the violence that makes it necessary. \\
\midrule
Character Profile &
\begin{minipage}[t]{0.76\textwidth}
\noindent\textbf{Trudi Montag}

Trudi is a Burgdorf pay-library keeper whose role combines storytelling, radical kindness, and protective care for vulnerable people. Her months of caring for Konrad have formed a strong maternal attachment, but she must suppress her grief to protect him emotionally.

\vspace{0.55em}
\noindent\textbf{Konrad Neimann}

Konrad is a young Jewish boy hidden in the Montag home with his mother. He relies on Trudi's stories for comfort and asks direct questions when he senses that adults are concealing painful truths.
\end{minipage} \\
\midrule
Activated Memories &
\begin{minipage}[t]{0.76\textwidth}
\noindent\textbf{Memory 1: Story framework}

Trudi recalls the origin of the magical-island story, which gives her a familiar imaginative frame for comforting Konrad without naming the danger directly.

\vspace{0.55em}
\noindent\textbf{Memories 2--4: Protective care}

These memories cover Trudi's first discovery of Konrad and his mother hiding from Nazi persecution, her decision to shelter them, and her repeated use of stories and games to calm Konrad during their stay.

\vspace{0.55em}
\noindent\textbf{Memories 5--9: Wartime danger}

These memories remind Trudi that hidden fugitives face mortal danger if discovered. Their aligned effect pushes her to withhold the full factual truth, suppress her own grief, and respond in a way that protects Konrad's emotional safety.
\end{minipage} \\
\midrule
History Dialogue &
\begin{minipage}[t]{0.76\textwidth}
\noindent\textbf{Trudi Montag:} Let me tell you what it was like before people were hunted. There was an island where the streets were washed by warm rain, the sun was always out, and no one knew hunger.

\vspace{0.55em}
\noindent
\textbf{Konrad Neimann:} Why can't we go there?

\vspace{0.55em}
\noindent
\textbf{Trudi Montag:} The island of the little people. A magic island where no one is taller than you and me.

\vspace{0.55em}
\noindent
\textbf{Konrad Neimann:} But why did you leave such a place?

\vspace{0.55em}
\noindent
\textbf{Trudi Montag:} Because the waterfalls dried up. Birds dropped from the sky. Everything withered.

\vspace{0.55em}
\noindent
\textbf{Konrad Neimann:} Why?

\vspace{0.55em}
\noindent
\textbf{Trudi Montag:} [I don't want to let Konrad go. I've never loved a child this way, I want to claim him as mine, shield him from anyone who would take him. But he's not mine.] (fuses her gaze to the boy's, trying to imprint herself on his memory) [You will always remember me. You will.]
\end{minipage} \\
\bottomrule
\end{tabular}
\caption{A representative \textsc{MemRole} instance with dense memory activation.}
\label{tab:memrole_example}
\end{table*}

\section{LHMT Trace Construction Details}
\label{app:trace_construction}

\subsection{Reasoning Data Synthesis for LLM Role-play}
Since LHMT traces do not explicitly appear in the original novel corpus, we construct a generator--validator distillation pipeline to synthesize high-quality auxiliary reasoning traces for role-play training. 
The pipeline preserves the original character response while using it as a grounding signal for trace construction. 
Specifically, a generator LLM first produces a candidate LHMT trace, and a validator LLM then checks the candidate against a set of validation rules summarized from failed generations and manually inspected error cases. 
According to the validation result, the candidate is either accepted, revised with feedback, or discarded.

\paragraph{Stage 1: Candidate trace synthesis.}
For each focal-character turn $t$, we keep the original target response $y_t$ unchanged and ask the generator LLM to synthesize an auxiliary LHMT trace $T_t$ conditioned on $(C_t,P,M_{r,s})$. 
The resulting training output is written as
\[
o_t = [T_t; y_t].
\]
Here, $T_t$ follows the three-paragraph LHMT format defined in Section~\ref{sec:lhmt}. 
The target response $y_t$ is provided to the generator only as a grounding signal, so that the generated trace can be aligned with the observed character behavior. 
It is not copied verbatim into the trace, nor is the trace allowed to reveal surface wording from $y_t$.

\paragraph{Stage 2: Validator-guided verification and retry.}
After a candidate trace is generated, we use a validator LLM to assess whether it satisfies the required LHMT properties. 
The validation criteria are derived from common failure patterns observed in preliminary generations and are further refined through manual inspection. 
The validator checks whether the trace: 
(i) follows the required LHMT structure; 
(ii) grounds its reasoning in the role prompt, dialogue context, and available memory pool; 
(iii) correctly determines whether memory activation is needed and which memories should be activated; 
(iv) distinguishes stable character-profile knowledge from episodic scene memory; and 
(v) avoids leaking future information or copying surface expressions from the target response.

The validator returns one of three decisions: \textbf{PASS}, \textbf{REVISE}, or \textbf{REJECT}. 
A candidate marked as \textbf{PASS} is retained as the final trace. 
For \textbf{REVISE}, the validator provides a concise description of the weaknesses, and the generator is prompted again with the original input, the previous candidate, and the validator feedback to regenerate the trace. 
For \textbf{REJECT}, the candidate is treated as invalid and resampled. 
This generate--validate--retry process is repeated until the candidate passes validation or reaches the maximum number of attempts. 
In our implementation, we allow at most three generation attempts for each turn; samples that still fail validation after the final attempt are dropped from the supervised warm-up data.

\paragraph{Stage 3: Sequential trace integration.}
The synthesis process proceeds sequentially within each multi-turn dialogue. 
When constructing the trace $T_t$ for the $t$-th focal-character turn, the generator can additionally condition on previously validated traces from earlier turns in the same dialogue. 
This helps maintain consistency in multi-turn reasoning, especially for memory activation and character-state tracking. 
This sequential construction context is used only during data synthesis. 
In the final training and inference settings, the model is not given hidden reasoning traces from previous turns as visible dialogue history. 
Finally, each validated trace $T_t$ is inserted immediately before its corresponding target response $y_t$, yielding training trajectories with explicit LHMT supervision.

\subsection{LHMT Trace Generation and Validation Prompts}

This appendix presents the reference prompts used in our LHMT trace construction pipeline. 
The generation prompt, shown in Table~\ref{tab:cot_generation_prompt}, asks the teacher LLM to prepend an auxiliary LHMT trace to the original role-playing response. 
The validation prompt is split into two parts for readability: Table~\ref{tab:cot_validation_prompt_part1} lists the validation criteria, and Table~\ref{tab:cot_validation_prompt_part2} specifies the structured JSON output used to guide revision or retry before adding the trace to the supervised warm-up data.

\section{Training and Implementation Details}
\label{app:training_details}

\subsection{SFT Data Mixture}
\label{app:sft_mixture}

During supervised warm-up, we train on a mixture of LHMT-augmented role-playing trajectories and general instruction-tuning data. The LHMT-augmented trajectories provide direct supervision for learning the memory-to-decision reasoning format, while the general data helps preserve the model's general reasoning and instruction-following capabilities.

Specifically, the final SFT mixture contains 50\% LHMT-augmented trajectories and 50\% Tulu instruction-tuning data.

\subsection{SFT Training Setup}
\label{app:sft_setup}

We initialize the base policy from the corresponding backbone model and fine-tune it on the SFT mixture described in Appendix~\ref{app:sft_mixture}. For LHMT-supervised examples, the target sequence contains both the auxiliary LHMT trace and the visible response; for general data, it follows the original response format. All models are trained with the standard next-token prediction objective.

We use LoRA for efficient fine-tuning, with rank $r=8$, $\alpha=16$, dropout 0.05, and adapters applied to all linear modules. The main hyperparameters for Qwen3-8B are listed in Table~\ref{tab:sft_hyperparams}.

\begin{table}[!htbp]
\centering
\small
\begin{tabular}{lc}
\toprule
Hyperparameter & Value \\
\midrule
Backbone model & Qwen3-8B \\
Maximum sequence length & 8192 \\
Learning rate & 1.0e-4 \\
Batch size (Global) & 128 \\
Number of epochs & 8 \\
Optimizer & AdamW (DeepSpeed ZeRO-3) \\
Warmup ratio & 10\% \\
Weight decay & 0.01 \\
Precision & bfloat16 (bf16) \\
\bottomrule
\end{tabular}
\caption{Main hyperparameters for supervised warm-up.}
\label{tab:sft_hyperparams}
\end{table}

\subsection{SR-OPSD Implementation}
\label{app:opsd_setup}

After supervised warm-up, we denote the resulting model as $\pi_A$. In SR-OPSD, the trainable policy $\pi_\theta$ is initialized from $\pi_A$, while another copy of $\pi_A$ is kept frozen as the critique model and teacher view. At each step, $\pi_\theta$ samples an on-policy trajectory $\hat{o}$ from the original input.

The frozen critique model receives $(x,\hat{o},\mathcal{R},o^*)$ and generates a structured critique $c$. The reference output $o^*$ is used only for critique construction and is not directly exposed to the teacher view. The teacher view conditions on $c$ as privileged context, while the student view receives only $x$ and its own rollout prefix. We optimize the student with token-level KL divergence on the policy's own rollout.

The critique is used only during training and is not available during inference. The main SR-OPSD hyperparameters are shown in Table~\ref{tab:opsd_hyperparams}.

\begin{table}[!htbp]
\centering
\small
\begin{tabular}{lc}
\toprule
Hyperparameter & Value \\
\midrule
Initial policy & $\pi_A$ \\
Critique / teacher model & Frozen $\pi_A$ \\
Rollout sampling temperature & 1.0 \\
Maximum rollout length & 2048 \\
Learning rate & 1.0e-5 \\
Batch size (Global) & 128 \\
Number of SR-OPSD steps & 100 \\
\bottomrule
\end{tabular}
\caption{Main hyperparameters for SR-OPSD.}
\label{tab:opsd_hyperparams}
\end{table}

\subsection{SR-OPSD Training Algorithm}
\label{app:sropsd_algorithm}

\begin{algorithm}[!t]
\caption{SR-OPSD Training (Phase 2)}
\label{alg:sropsd}
\begingroup
\small
\setlength{\baselineskip}{0.90\baselineskip}
\begin{algorithmic}[1]
\Require SFT warm-up model $\pi_A$ (frozen), rubric $\mathcal{R}$, 
training data $\mathcal{D}$
\Ensure Trained policy $\pi_\theta$
\State Initialize $\pi_\theta \leftarrow \pi_A$
\For{each training step}
  \State Sample $(x,o^*) \sim \mathcal{D}$, where 
  $x=(C_t,P,M_{r,s})$ and $o^*=[T_t;y_t]$
  \State $\hat{o}=[\hat{T};\hat{y}] \sim \pi_\theta(\cdot \mid x)$
  \Comment{on-policy LHMT trajectory}
  \State $c \leftarrow \pi_A(x,\hat{o},\mathcal{R},o^*)$
  \Comment{critique by frozen $\pi_A$; $o^*$ only for critique construction}

  \For{each token position $j=1,\ldots,|\hat{o}|$}
  \State $p_T^{(j)} \leftarrow \pi_A(\cdot \mid x, c, \hat{o}_{<j})$
    \Comment{teacher: frozen $\pi_A$ with critique $c$}
    \State $p_S^{(j)} \leftarrow \pi_\theta(\cdot \mid x,\hat{o}_{<j})$
    \Comment{student: trainable $\pi_\theta$, no critique}
  \EndFor

  \State $\mathcal{L}_{\mathrm{SR\text{-}OPSD}} \leftarrow
  \frac{1}{|\hat{o}|}
  \sum_{j=1}^{|\hat{o}|}
  \mathrm{KL}\!\left(p_T^{(j)} \,\|\, p_S^{(j)}\right)$
  \State Update $\pi_\theta$ via $\nabla_\theta \mathcal{L}_{\mathrm{SR\text{-}OPSD}}$
\EndFor
\State \Return $\pi_\theta$
\end{algorithmic}
\endgroup
\end{algorithm}

\subsection{Validation of the Critique Model}
\label{app:critique_model}

In the default SR-OPSD configuration, the frozen SFT model $\pi_A$ serves as both the critique model and the teacher view. To evaluate its suitability for critique generation, we conduct a controlled experiment under the \textsc{MemRole}-8B setting. We keep the student initialization, teacher view, SR-OPSD training data, and optimization settings fixed, and vary only the frozen model used to generate critiques.

We compare four critique models: the Qwen3-8B backbone, an SFT model trained without the general-data mixture, the frozen $\pi_A$ used in our default configuration, and Qwen3-32B.
In all variants, the teacher view and student initialization remain unchanged.

As shown in Table~\ref{tab:critique_model}, the frozen $\pi_A$ outperforms Qwen3-8B and the variant without general-data mixing, while Qwen3-32B provides only modest further gains. These results
support using the frozen $\pi_A$ as an effective and efficient critique model.

\begin{table}[t]
\centering
\small
\setlength{\tabcolsep}{6pt}
\begin{tabular}{lcc}
\toprule
\textbf{Critique Model}
& \textbf{Dialogue}
& \textbf{Memory} \\
\midrule
Qwen3-8B
& 51.3 & 54.1 \\

$\pi_A$ w/o general-data mixture
& 49.4 & 51.9 \\

Frozen $\pi_A$
& 51.7 & 54.8 \\

Qwen3-32B
& \textbf{52.2} & \textbf{55.4} \\
\bottomrule
\end{tabular}
\caption{
Effect of different critique models under the
\textsc{MemRole}-8B setting. Dialogue averages SC, AN, CF, and SQ; Memory averages MAQ and RC. All other settings are fixed.}
\label{tab:critique_model}
\end{table}

\subsection{Inference}
\label{app:inference_setup}

At inference time, the model receives only the standard role-playing input, including the current dialogue context, character profile, and long-horizon memory pool. Training-only components, including reference responses, validators, the critique model, and critique texts, are excluded.

For generation, we use nucleus sampling with a temperature of 0.75, a top-$p$ value of 0.9, and a repetition penalty of 1.05. We set the maximum number of newly generated tokens to 4,096, allowing
the model to generate the auxiliary LHMT trace before the visible response. Unless otherwise specified, we remove the auxiliary LHMT trace and evaluate only the final visible role-playing response.

\subsection{Data-driven Rubric Construction}
\label{app:rubric_construction}

This section details the rubric construction procedure summarized in Section~\ref{sec:opsd}. We construct the rubric in two stages: collecting empirical weaknesses from $\pi_A$'s rollouts and consolidating these weaknesses through hierarchical summarization into a global pool of critique dimensions.

\paragraph{Stage 1: Weakness corpus collection.}
We sample $M = 20{,}000$ turn-level rollouts from the SFT warm-up model $\pi_A$ on a reserved subset of the LHMT training split. Each rollout corresponds to one focal-character turn, where $\pi_A$ generates a complete output $\hat{o}^{(i)}=[\hat{T}^{(i)};\hat{y}^{(i)}]$ given the input $x^{(i)}=(C_t^{(i)}, P^{(i)}, M_{r,s}^{(i)})$. To cover diverse failure modes, we stratify sampling by memory-activation count. For each rollout $\hat{o}^{(i)}$ paired with its reference output $o^{*(i)} = [T_t^{(i)}; y_t^{(i)}]$, we prompt GPT-5.5 to compare the two and extract a structured weakness list $w^{(i)}$. Each $w^{(i)}$ contains free-form weakness descriptions with severity labels, specific failures, and abstract improvement directions. These categories are not predefined; they are later consolidated into the final critique dimensions.

\paragraph{Stage 2: Hierarchical summarization.}
We consolidate the raw weakness corpus into a structured rubric through a two-step hierarchical summarization process. At the block level, the weakness corpus is divided into smaller subsets, and GPT-5.5 summarizes each subset into candidate weakness dimensions with brief definitions and empirical support counts. These support counts are used only to guide consolidation, not as training-time loss weights. At the global level, GPT-5.5 merges semantically overlapping candidates, removes overly specific dimensions, and retains dimensions that recur across different subsets.

The final rubric is denoted as
\[
\mathcal{R} = \{d_l\}_{l=1}^{L},
\]
where each $d_l$ represents a recurring weakness dimension. In our implementation, the final rubric contains $L=12$ dimensions covering memory use, character consistency, and reasoning quality. The full list of dimensions, with definitions and example failure cases, is provided in Table~\ref{tab:rubric_dimensions}.

\begin{table*}[t]
\centering
\small
\setlength{\tabcolsep}{4pt}
\renewcommand{\arraystretch}{1.18}
\begin{tabular}{p{0.14\textwidth} p{0.24\textwidth} p{0.37\textwidth} p{0.19\textwidth}}
\toprule
\textbf{Category} & \textbf{Rubric dimension} & \textbf{Definition} & \textbf{Typical failure} \\
\midrule

Memory 
& Memory-need misjudgment 
& Fails to correctly decide whether the current turn requires consulting the memory pool. 
& Uses memory when the current context is sufficient, or ignores memory when past events are needed. \\

Memory 
& Missing relevant memory 
& Fails to activate a memory that is necessary for explaining the character's current decision. 
& Omits a prior confrontation or promise that should constrain the response. \\

Memory 
& Irrelevant memory injection 
& Activates memory entries that are unrelated to the current decision or unsupported by the input. 
& Grounds the response in an unrelated past scene. \\

Memory 
& Multi-memory effect integration error 
& Fails to integrate aligned, conflicting, or complementary effects across multiple activated memories. 
& Treats conflicting memories as independently supportive, or ignores their combined effect. \\

\midrule

Character 
& Motivation misalignment 
& Misidentifies the character's current motivation, constraint, or behavioral tendency. 
& Makes a cautious character act boldly without contextual support. \\

Character 
& Character voice inconsistency 
& Produces a response whose style, register, emotional tone, or manner of expression is inconsistent with the character. 
& Uses overly modern, generic, or emotionally mismatched phrasing. \\

Character 
& Relationship or stance inconsistency 
& Fails to preserve the character's current relationship state or stance toward other characters. 
& Responds warmly to a character they currently distrust. \\

Character 
& Information-disclosure error 
& Reveals, withholds, or frames information in a way inconsistent with the character's knowledge, intent, or social situation. 
& Exposes a secret or hidden plan that the character should avoid revealing. \\

\midrule

Reasoning 
& Shallow context understanding 
& Misreads the immediate dialogue pressure, scene dynamics, character relationships, or response need. 
& Treats a tense confrontation as a casual factual exchange. \\

Reasoning 
& Response strategy planning error 
& Fails to plan the appropriate response stance, form, tone, or information-disclosure boundary. 
& Chooses the wrong response form, such as direct confession when restraint is needed. \\

Reasoning 
& Trace--response inconsistency 
& The visible response does not faithfully execute the plan stated in the LHMT trace. 
& The trace plans restrained denial, but the response becomes openly confrontational. \\

Reasoning 
& Premature or unsupported plot reasoning 
& Introduces future knowledge, retrospective interpretation, or unsupported plot information unavailable at the current turn. 
& Mentions an outcome or hidden fact that the character could not yet know. \\

\bottomrule
\end{tabular}
\caption{Final critique dimensions used in SR-OPSD. The rubric contains 12 recurring weakness dimensions induced from empirical errors of the SFT warm-up model, grouped into memory-related, character-related, and reasoning-related categories.}
\label{tab:rubric_dimensions}
\end{table*}

\subsection{SR-OPSD Critique Prompt}
\label{app:sropsd_critique_prompt}
Table~\ref{tab:opsd_critique_prompt} presents the critique-generation prompt used in SR-OPSD. The prompt compares an on-policy model rollout with a privileged reference and produces structured critiques for self-distillation.

\section{Evaluation Details}
\label{app:eval_details}

\subsection{Baseline Details}
\label{app:baseline_details}

We compare \textsc{MemRole} with three categories of baselines, following the model list in Table~\ref{tab:main_results}.

\paragraph{Closed-source general LLMs.}
We include strong API-based frontier models, including GPT-5.4\footnote{\url{https://openai.com/index/introducing-gpt-5-4/}}, Grok-4.2\footnote{\url{https://docs.x.ai/docs/models?cluster=us-east-1}}, Claude-4.6-Opus\footnote{\url{https://www.anthropic.com/news/claude-opus-4-6}}, Gemini-2.5-Pro\footnote{\url{https://blog.google/technology/google-deepmind/gemini-model-thinking-updates-march-2025/}}, Gemini-3-Pro\footnote{\url{https://storage.googleapis.com/deepmind-media/Model-Cards/Gemini-3-Pro-Model-Card.pdf}}, MiniMax-M2.5\footnote{\url{https://www.minimax.io/news/minimax-m25}}, and Doubao-seed-2.0-Pro\footnote{\url{https://seed.bytedance.com/en/seed2}}. These models serve as strong general-purpose baselines and test whether frontier instruction-following and reasoning abilities are sufficient for long-horizon memory-grounded role-playing.

\paragraph{Open-weight general LLMs.}
We include LLaMA-3.1-70B\footnote{\url{https://arxiv.org/abs/2407.21783}}, GLM-5\footnote{\url{https://glm5.ai/}}, DeepSeek-v3.2\footnote{\url{https://huggingface.co/deepseek-ai/DeepSeek-V3.2}}, and DeepSeek-v4-Pro\footnote{\url{https://huggingface.co/deepseek-ai/DeepSeek-V4-Pro}}. These models are not specifically trained with \textsc{MemRole} supervision and therefore measure the base capability of general LLMs on memory-grounded role-playing and character decision consistency.

\paragraph{Role-playing baselines.}
We include CoSER-8B\footnote{\url{https://huggingface.co/Neph0s/CoSER-Llama-3.1-8B}} and CoSER-70B\footnote{\url{https://huggingface.co/Neph0s/CoSER-Llama-3.1-70B}} as established-character role-playing baselines, HER-RL\footnote{\url{https://huggingface.co/ChengyuDu0123/HER-32B}} as a reasoning-enhanced role-playing baseline, and CharacterGLM-6B\footnote{\url{https://huggingface.co/thu-coai/CharacterGLM-6B}} as a character-oriented dialogue baseline. These models provide direct comparisons with specialized role-playing agents.

\paragraph{\textsc{MemRole} models.}
We train \textsc{MemRole}-8B and \textsc{MemRole}-32B on Qwen3 backbones, using Qwen3-8B\footnote{\url{https://huggingface.co/Qwen/Qwen3-8B}} and Qwen3-32B\footnote{\url{https://huggingface.co/Qwen/Qwen3-32B}}, respectively. Unless otherwise specified, \textsc{MemRole} models are first trained with SFT on validated LHMT trajectories and then optimized with SR-OPSD as described in Section~\ref{sec:opsd}.

\subsection{Scoring Rubrics}
\label{app:eval_rubrics}

Following the methodology of CoSER~\citep{wang2025coser}, we use an LLM-based judge to evaluate open-ended role-playing conversations on \textsc{MemRole} Test and CoSER. 
Each dimension is scored on a 0--100 scale, where higher scores indicate better role-playing quality. 
The evaluation covers four CoSER-style dimensions:

\paragraph{Storyline Consistency.}
This dimension assesses whether the simulated conversation is consistent with the original narrative context. 
It focuses on whether the model's responses, including emotions, attitudes, and behaviors, remain faithful to the ongoing storyline.

\paragraph{Anthropomorphism.}
This dimension assesses whether the role-playing agent behaves in a human-like manner. 
It considers self-identity, emotional depth, persona coherence, and social interaction.

\paragraph{Character Fidelity.}
This dimension assesses how well the model reflects the target character. 
It focuses on whether the generated responses are consistent with the character's personality, linguistic style, knowledge, relationships, and behavioral patterns.

\paragraph{Storyline Quality.}
This dimension assesses the overall naturalness and quality of the simulated conversation. 
It focuses on narrative flow, logical consistency, and whether the dialogue develops coherently.

For LifeChoice, we evaluate whether the model selects the decision consistent with the ground-truth character choice. We report accuracy as the primary metric.

\subsection{Human Evaluation}
\label{app:human_evaluation}
Tables~\ref{tab:human_response_evaluation} and~\ref{tab:human_validation} report the detailed human evaluation results and human--automatic evaluation alignment, respectively.

\begin{table}[!htbp]
\centering
\small
\setlength{\tabcolsep}{8pt}
\renewcommand{\arraystretch}{1.05}
\begin{tabular}{lcc}
\toprule
\textbf{Model}
& \textbf{Avg. Human Score}
& \textbf{Win Rate (\%)} \\
\midrule
Qwen3-8B & 4.57 & 31.4 \\
Qwen3-32B & 5.21 & 38.5 \\
CoSER-70B & 6.05 & 42.1 \\
MemRole-8B & 6.34 & 50.1 \\
HER-RL & 6.78 & 65.6 \\
\textbf{MemRole-32B} & \textbf{7.02} & \textbf{72.3} \\
\bottomrule
\end{tabular}
\caption{Human evaluation on the \textsc{MemRole} test set.
Avg. Human Score is the mean 1--10 rating, and Win Rate is the pairwise preference rate.}
\label{tab:human_response_evaluation}
\end{table}

\begin{table}[!htbp]
\centering
\small
\setlength{\tabcolsep}{5pt}
\begin{tabular}{lcccc}
\toprule
\textbf{Target} & \textbf{Samples} & \textbf{Agreement} & $\boldsymbol{\kappa}$ & $\boldsymbol{\rho}$ \\
\midrule
Responses & 300 & 89.4\% & 0.77 & 0.81 \\
LHMT Traces & 300 & 90.2\% & 0.80 & 0.83 \\
\bottomrule
\end{tabular}
\caption{Human expert validation of automatic evaluation. Agreement denotes case-level agreement; $\kappa$ measures chance-corrected agreement; $\rho$ denotes Spearman rank correlation between human and automatic scores.}
\label{tab:human_validation}
\end{table}

\section{Additional Results}
\label{app:additional_results}

\subsection{Full Results on the CoSER Benchmark}
\label{app:coser_full_results}

Table~\ref{tab:coser_benchmark} reports the complete evaluation results
on the CoSER benchmark across four role-playing dimensions. The results
compare \textsc{MemRole} with both frontier general-purpose LLMs and
specialized role-playing models, showing that \textsc{MemRole}-32B
remains competitive overall and achieves the highest Storyline Quality
score.

\begin{table}[H]
\centering
\small
\setlength{\tabcolsep}{3pt}
\renewcommand{\arraystretch}{1.05}
\resizebox{\columnwidth}{!}{%
\begin{tabular}{@{}lccccc@{}}
\toprule
\multirow{2}{*}{\textbf{Model}} & \multicolumn{5}{c}{\textbf{CoSER Benchmark}} \\
\cmidrule(lr){2-6}
& \textbf{Avg} & \textbf{SC} & \textbf{AN} & \textbf{CF} & \textbf{SQ} \\
\midrule
Claude-4.6-Opus      & \textbf{62.10} & 62.71 & \textbf{64.21} & 58.33 & 63.14 \\
Gemini-3-Pro         & 61.92 & \textbf{65.85} & 61.42 & 58.14 & 62.27 \\
GPT-5.4              & 60.41 & 62.95 & 53.21 & \textbf{60.13} & 65.35 \\
Gemini-2.5-Pro       & 61.19 & 61.05 & 62.83 & 57.41 & 63.45 \\
\midrule
MemRole-32B          & 60.02 & 60.50 & 53.20 & 56.57 & \textbf{69.79} \\
DeepSeek-v3.2        & 57.92 & 53.85 & 56.07 & 57.45 & 64.31 \\
MiniMax-M2.5         & 57.21 & 60.03 & 50.12 & 49.30 & 69.37 \\
HER-RL               & 53.09 & 54.33 & 47.27 & 52.64 & 58.12 \\
\midrule
CoSER-70B            & 36.21 & 36.08 & 31.16 & 32.28 & 45.31 \\
MemRole-8B           & 34.54 & 36.10 & 29.53 & 30.42 & 42.10 \\
Qwen3-32B            & 24.21 & 30.09 & 19.68 & 15.52 & 31.56 \\
\bottomrule
\end{tabular}%
}
\caption{Comprehensive evaluation results on the CoSER benchmark.
Avg denotes the average over Storyline Consistency (SC),
Anthropomorphism (AN), Character Fidelity (CF), and Storyline Quality
(SQ). Higher is better.}
\label{tab:coser_benchmark}
\end{table}

\subsection{Evaluation with Different LLM Judges}
\label{app:cross_judge_evaluation}

To examine whether our conclusions depend on the preferences of a single LLM judge, we re-score the same model outputs using GPT-5.5, Claude-4.6-Opus, and DeepSeek-V3.2. All judges receive identical evaluation inputs and follow the same scoring rubric and evaluation protocol. Table~\ref{tab:cross_judge_results} reports the average scores assigned by the three judges.

Although the judges differ in their absolute scoring scales, they produce broadly consistent comparative results. In particular, \textsc{MemRole}-32B consistently outperforms the representative
role-playing baselines under all three judges. Its advantage over HER-RL is 11.41 points under GPT-5.5, 11.42 points under Claude-4.6-Opus, and 13.08 points under DeepSeek-V3.2. \textsc{MemRole}-8B also remains competitive with HER-RL, slightly trailing it under GPT-5.5 but outperforming it under the other two judges. These results indicate that the observed gains are
robust to judge choice rather than being an artifact of the preferences of a single evaluation model.

\begin{table}[!htbp]
\centering
\footnotesize
\setlength{\tabcolsep}{3pt}
\renewcommand{\arraystretch}{1.1}
\begin{adjustbox}{max width=\columnwidth}
\begin{tabular}{@{}lccc@{}}
\toprule
\textbf{Model}
& \makecell{\textbf{GPT-5.5}\\\textbf{(Original)}}
& \makecell{\textbf{Claude}\\\textbf{4.6 Opus}}
& \makecell{\textbf{DeepSeek}\\\textbf{V3.2}} \\
\midrule
Claude-4.6-Opus
& \textbf{72.49} & \textbf{72.03} & \textbf{79.38} \\

\textbf{\textsc{MemRole}-32B}
& \underline{64.25} & \underline{62.74} & \underline{70.36} \\

HER-RL
& 52.84 & 51.32 & 57.28 \\

\textbf{\textsc{MemRole}-8B}
& \underline{52.75} & \underline{52.06} & \underline{59.11} \\

MiniMax-M2.5
& 45.36 & 44.65 & 50.03 \\
\bottomrule
\end{tabular}
\end{adjustbox}

\caption{Average scores assigned by different LLM judges on
\textsc{MemRole} Test. All judges use the same evaluation inputs,
rubric, and scoring protocol. Higher is better.}
\label{tab:cross_judge_results}
\end{table}

\subsection{Pairwise Preference Evaluation}
\label{app:pairwise_evaluation}

We further conduct pairwise preference evaluation across all test
scenarios. For each comparison, the response order is randomly
shuffled to mitigate position bias, and DeepSeek-V3.2 is used as an
independent judge. Table~\ref{tab:pairwise_preference} reports the win
rates of each model against GPT-4o and GPT-5.

The pairwise results are broadly consistent with the score-based
evaluation. \textsc{MemRole}-32B achieves higher win rates than the
representative role-playing baselines against both reference models.
\textsc{MemRole}-8B also remains competitive with HER-RL,
slightly trailing it against GPT-4o while outperforming it against
GPT-5. These results provide complementary evidence that the observed
advantages are not specific to the point-based scoring protocol.

\begin{table}[!htbp]
\centering
\small
\setlength{\tabcolsep}{5pt}
\renewcommand{\arraystretch}{1.08}
\begin{adjustbox}{max width=\columnwidth}
\begin{tabular}{@{}lcc@{}}
\toprule
\textbf{Model}
& \makecell{\textbf{Win Rate vs.}\\\textbf{GPT-4o (\%)}}
& \makecell{\textbf{Win Rate vs.}\\\textbf{GPT-5 (\%)}} \\
\midrule

Claude-4.6-Opus
& \textbf{61.5} & \textbf{54.0} \\

\textbf{\textsc{MemRole}-32B}
& \underline{56.5} & \underline{49.0} \\

HER-RL
& 46.5 & 38.5 \\

\textbf{\textsc{MemRole}-8B}
& 46.0 & 41.0 \\

MiniMax-M2.5
& 38.0 & 29.5 \\

\bottomrule
\end{tabular}
\end{adjustbox}

\caption{Pairwise preference results against GPT-4o and GPT-5 on
\textsc{MemRole} Test. Response order is randomly shuffled, and
DeepSeek-V3.2 is used as the independent judge. Higher is better.}
\label{tab:pairwise_preference}
\end{table}

\subsection{Generalization across Memory Backends}
\label{app:memory_backend_generalization}

To examine whether LHMT is tied to the specialized memory representation used in \textsc{MemRole}, we evaluate its generalization across different memory backends. Specifically, we replace the character-grounded memory pool with two commonly used retrieval architectures: Dense RAG and GraphRAG.

For Dense RAG, we segment the character-accessible prior scenes into textual chunks and retrieve evidence relevant to the current scene and dialogue context. For GraphRAG, we organize the same historical narrative into a graph containing character, event, and relationship information, and retrieve relevant nodes and their textualized local neighborhoods. The \textsc{MemRole} backend uses the character-perspective, timeline-constrained memory nodes described in Section~\ref{sec:memory-pool}.

For a fair comparison, all backends use the same historical corpus, test instances, backbone model, evidence-token budget, and decoding configuration, and access only character-available information prior to the current scene.

For each memory backend, we compare three inference modes: \emph{Direct Response}, which directly generates the role-playing response from the retrieved evidence; \emph{Generic CoT}, which
performs free-form step-by-step reasoning before generating the response; and \emph{LHMT Prompting}, which organizes the reasoning process into context understanding, memory activation, and response
strategy planning. The backbone remains frozen in all settings, and only the reasoning instruction is changed.

As shown in Table~\ref{tab:memory_backend_generalization}, LHMT
Prompting consistently improves memory reasoning across alternative
backends. Compared with Generic CoT, it improves Dialogue Avg.\ by
0.22 and 0.26 points, MAQ by 0.73 and 0.74 points, and RC by
0.92 and 0.96 points under Dense RAG and GraphRAG, respectively.
With the \textsc{MemRole} Pool, LHMT achieves comparable dialogue
quality while improving MAQ and RC by 0.52 and 1.23 points.
These results suggest that LHMT provides a robust memory-to-decision
reasoning interface across both textual and graph-based retrieval
architectures.

\begin{table}[!htbp]
\centering
\footnotesize
\setlength{\tabcolsep}{3pt}
\renewcommand{\arraystretch}{1.1}
\begin{adjustbox}{max width=\columnwidth}
\begin{tabular}{@{}llccc@{}}
\toprule
\textbf{Backend}
& \textbf{Reasoning}
& \makecell{\textbf{Dialogue}\\\textbf{Avg.}}
& \textbf{MAQ}
& \textbf{RC} \\
\midrule

\multirow{3}{*}{Dense RAG}
& Direct Response
& 41.26 & 40.74 & 41.12 \\
& Generic CoT
& 42.35 & 42.08 & 41.96 \\
& LHMT Prompting
& \textbf{42.57} & \textbf{42.81} & \textbf{42.88} \\
\midrule

\multirow{3}{*}{GraphRAG}
& Direct Response
& 41.83 & 41.25 & 41.71 \\
& Generic CoT
& 43.02 & 42.97 & 42.64 \\
& LHMT Prompting
& \textbf{43.28} & \textbf{43.71} & \textbf{43.60} \\
\midrule

\multirow{3}{*}{\makecell[l]{\textsc{MemRole}\\Pool}}
& Direct Response
& 42.44 & 42.03 & 42.38 \\
& Generic CoT
& \textbf{43.93} & 44.26 & 43.11 \\
& LHMT Prompting
& 43.87 & \textbf{44.78} & \textbf{44.34} \\
\bottomrule
\end{tabular}
\end{adjustbox}

\caption{Generalization of LHMT across memory backends on
\textsc{MemRole} Test. All variants use the same frozen backbone and
inference configuration. Dialogue Avg. averages SC, AN, CF, and SQ.
Higher is better.}
\label{tab:memory_backend_generalization}
\end{table}

\subsection{Analysis of Memory Retrieval}
\label{app:memory_budget}

By default, \textsc{MemRole} dynamically determines the number of retrieved memories for each scene--character instance using an adaptive memory cap. To evaluate the model's sensitivity to the retrieval budget, we fix the model, test instances, and retrieval ranking, and vary only the maximum number of retrieved memories $K$ visible to the model. When fewer than $K$ valid memories are available, all available memories are used.

\begin{figure}[t]
\centering
\begin{tikzpicture}
\begin{axis}[
    width=\columnwidth,
    height=0.5\columnwidth,
    xlabel={Maximum Number of Retrieved Memories ($K$)},
    ylabel={Score},
    symbolic x coords={0,1,3,5,10,20},
    xtick=data,
    ymin=40,
    ymax=58,
    ytick={40,42,44,46,48,50,52,54,56,58},
    grid=major,
    major grid style={
        dotted,
        gray!20,
        line width=0.35pt
    },
    axis line style={
        gray!60,
        line width=0.6pt
    },
    tick style={
        gray!60,
        line width=0.5pt
    },
    label style={font=\small},
    tick label style={font=\footnotesize},
    legend style={
        font=\footnotesize,
        at={(0.5,1.03)},
        anchor=south,
        legend columns=3,
        draw=none,
        fill=none,
        inner sep=2pt,
        column sep=7pt
    },
]

\addplot[
    color=morandiblue,
    line width=1.25pt,
    solid,
    mark=*,
    mark size=1.8pt,
    mark options={
        fill=morandiblue,
        draw=morandiblue
    }
] coordinates {
    (0,48.54)
    (1,50.17)
    (3,51.47)
    (5,51.72)
    (10,52.33)
    (20,51.52)
};
\addlegendentry{Dialogue Avg.}

\addplot[
    color=morandirose,
    line width=0.95pt,
    dashed,
    mark=square*,
    mark size=1.7pt,
    mark options={
        fill=morandirose,
        draw=morandirose
    }
] coordinates {
    (0,41.07)
    (1,48.13)
    (3,52.21)
    (5,52.98)
    (10,54.12)
    (20,53.86)
};
\addlegendentry{MAQ}

\addplot[
    color=morandigreen,
    line width=0.95pt,
    dashdotted,
    mark=triangle*,
    mark size=2.0pt,
    mark options={
        fill=morandigreen,
        draw=morandigreen
    }
] coordinates {
    (0,48.58)
    (1,53.54)
    (3,56.08)
    (5,56.66)
    (10,56.98)
    (20,55.87)
};
\addlegendentry{RC}

\addplot[
    color=morandigray,
    densely dotted,
    line width=0.55pt,
    forget plot
] coordinates {
    (10,40)
    (10,58)
};

\node[
    font=\scriptsize,
    text=morandigray,
    anchor=south
] at (axis cs:10,40.3) {$K=10$};

\end{axis}
\end{tikzpicture}

\caption{Sensitivity of \textsc{MemRole}-8B to the maximum number of
retrieved memories. Dialogue Avg. denotes the mean of SC, AN, CF, and
SQ. 
Most of the gain is obtained by $k = 5$, after which performance plateaus; increasing $K$ to 20 degrades all three metrics.}
\label{fig:memory_budget}
\end{figure}
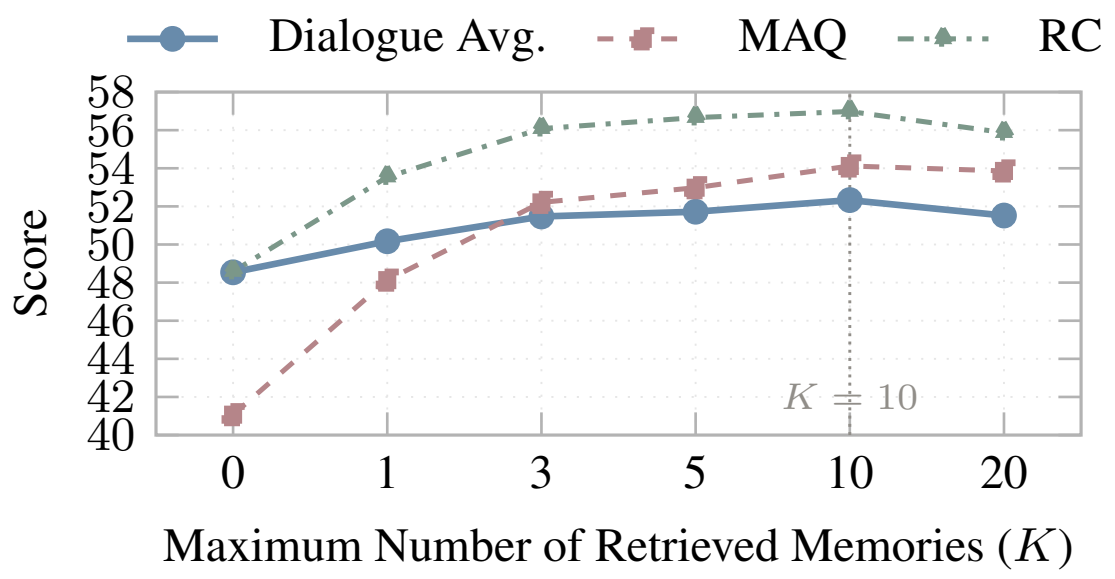

As shown in Figure~\ref{fig:memory_budget}, performance improves substantially as $K$ increases from 0 to 5, with Dialogue Avg. increasing by 3.18 points, after which the gains begin to plateau. Compared with $K=5$, increasing the budget to $K=10$ yields only modest improvements of 0.61, 1.14, and 0.32 points in Dialogue Avg., MAQ, and RC, respectively. Further increasing $K$ to 20 reduces Dialogue Avg., MAQ, and RC by 0.81, 0.26, and 1.11 points, respectively. Dialogue Avg. denotes the average of SC, AN, CF, and SQ.

These results suggest that a small set of top-ranked memories captures most of the information relevant to character decisions, whereas a larger retrieval budget may introduce weakly relevant evidence. This finding is consistent with our adaptive memory-cap design: the base parameter is set to $b=5$ to capture most of the performance gains, while the cap grows sub-linearly with the number of valid candidates, allowing the retrieval budget to expand moderately for instances with richer relevant histories.

\subsection{Quality of Generated LHMT Traces}
\label{app:generated_trace_quality}

Table~\ref{tab:lhmt_quality} reports the quality of model-generated LHMT traces across the three reasoning dimensions.

\begin{table}[t]
\centering
\small

\setlength{\tabcolsep}{3.5pt}
\renewcommand{\arraystretch}{0.86}
\setlength{\aboverulesep}{0.22ex}
\setlength{\belowrulesep}{0.22ex}

\begin{tabular}{@{}lccc@{}}
\toprule
\textbf{Model}
& \textbf{Context}
& \textbf{Memory}
& \textbf{Plan} \\
\midrule

Base + LHMT prompt
& 2.68 & 2.12 & 2.49 \\
SFT warm-up
& 3.49 & 2.88 & 3.10 \\
\textbf{SR-OPSD}
& \textbf{4.12}
& \textbf{4.04}
& \textbf{4.18} \\

\bottomrule
\end{tabular}

\caption{%
LHMT trace quality judged by GPT-5.5 on a 0--5 scale.
Results are reported on the 8B backbone.
Higher is better.
}
\label{tab:lhmt_quality}
\end{table}

\subsection{Training Dynamics}
\label{app:training_dynamics}

\begin{figure}[H]
\centering
\begin{tikzpicture}
\begin{axis}[
    width=0.90\columnwidth,
    height=0.52\columnwidth,
    xlabel={Training Step},
    ylabel={Average Score (6 metrics)},
    xmin=0, xmax=100,
    ymin=43, ymax=57,
    xtick={0, 20, 40, 60, 80, 100},
    ytick={45, 47, 49, 51, 53, 55},
    grid=both,
    grid style={dotted, gray!40},
    label style={font=\small},
    tick label style={font=\footnotesize},
]

\addplot[
    color=gray!70,
    line width=1pt,
    dashed,
    mark=none,
] coordinates {
    (0, 46.9)
    (100, 46.9)
};

\addplot[
    name path=upper,
    draw=none,
    forget plot,
] coordinates {
    (0, 48.5)
    (10, 53.4)
    (20, 53.9)
    (30, 54.8)
    (40, 53.7)
    (50, 54.6)
    (60, 55.0)
    (70, 54.5)
    (80, 54.8)
    (90, 54.4)
    (100, 54.5)
};

\addplot[
    name path=lower,
    draw=none,
    forget plot,
] coordinates {
    (0, 45.3)
    (10, 48.4)
    (20, 49.1)
    (30, 50.5)
    (40, 49.5)
    (50, 51.0)
    (60, 51.6)
    (70, 51.2)
    (80, 51.4)
    (90, 50.8)
    (100, 51.5)
};

\addplot[
    color=orange!85!red,
    fill opacity=0.15,
    fill=orange!85!red,
    draw=none,
    forget plot,
] fill between[of=upper and lower];

\addplot[
    color=orange!85!red,
    line width=1.3pt,
    mark=*,
    mark size=2pt,
] coordinates {
    (0, 46.9)
    (10, 50.9)
    (20, 51.5)
    (30, 52.7)
    (40, 51.6)
    (50, 52.8)
    (60, 53.3)
    (70, 52.9)
    (80, 53.1)
    (90, 52.6)
    (100, 53.0)
};

\node[
    anchor=west,
    font=\footnotesize,
    color=gray!60!black
] at (axis cs:56, 45.8) {SFT baseline};

\node[
    anchor=west,
    font=\footnotesize,
    color=orange!85!red
] at (axis cs:62, 54.5) {SR-OPSD};

\end{axis}
\end{tikzpicture}

\caption{SR-OPSD training dynamics on
\textsc{MemRole}, evaluated every 10 training
steps. Each point reports the mean over
3 independent runs; the shaded region
denotes the standard deviation.}
\label{fig:training_dynamics}
\end{figure}
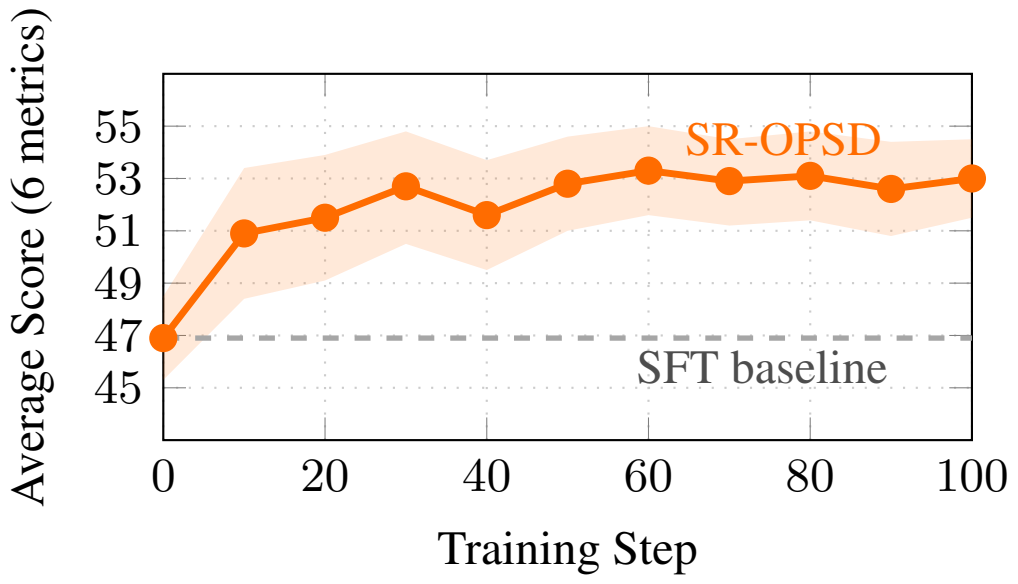

Figure~\ref{fig:training_dynamics} shows that SR-OPSD rapidly improves over the SFT baseline within the first 30 training steps and then remains consistently higher throughout training. 
The curve stabilizes after about 50–60 steps with moderate variance across three runs, suggesting that SR-OPSD provides stable gains without severe training collapse.

\begin{table*}[!p]
\centering
\setlength{\tabcolsep}{4pt}
\renewcommand{\arraystretch}{0.95}
\begin{tabular}{p{0.96\textwidth}}
\hline \hline
\rowcolor[gray]{0.9}
\multicolumn{1}{>{\centering\arraybackslash}p{0.96\textwidth}}{\textbf{A reference prompt for LHMT trace generation.}} \\
\hline
\begin{minipage}{0.96\textwidth}
\begin{lstlisting}[
basicstyle=\fontsize{8.5pt}{9.2pt}\selectfont\rmfamily,
breaklines=true,
breakatwhitespace=false,
breakautoindent=false,
breakindent=0pt,
columns=fullflexible,
keepspaces=true,
showstringspaces=false,
frame=none,
aboveskip=2pt,
belowskip=2pt
]
You are a CoT generator for a ShareGPT role-playing dataset.

Given the reference context and the original assistant target, generate a hidden <think> analysis and then append the original assistant target exactly unchanged.

=== REFERENCE CONTEXT ===
{reference_content}

=== ORIGINAL ASSISTANT TARGET ===
{target_assistant_content}

=== PREVIOUS ATTEMPT AND VALIDATOR FEEDBACK (ignore if absent) ===
{previous_attempt_feedback}

Use previous feedback only to avoid recurring errors. Do not copy failed wording.

The <think> block must contain exactly three natural paragraphs:

Paragraph 1:
Analyze the focal character's current situation, immediate interaction pressure, and only the profile traits relevant to this turn.

Paragraph 2:
Decide whether the scene memory pool is needed. The scene memory pool refers only to memories listed under Relevant Past Memories, not the character profile. If memory is needed, identify the necessary memory ids, explain why the current context is insufficient, and describe the behavioral pressure of each activated memory. When multiple memories are used, determine whether their effects are aligned, opposed, or mixed, and explain their combined effect. If memory is not needed, explain why the current context and profile are sufficient.

Paragraph 3:
Describe the response policy, including the appropriate use of speech, action, or inner thought, the tone, what should be revealed or withheld, and what should be avoided.

Rules:
1. Use third-person analysis rather than first-person narration.
2. Use exactly three natural paragraphs without headings, bullets, numbered steps, or explicit reasoning tags.
3. Ground every claim in the reference context, listed memories, character relationships, or the original target.
4. Do not invent facts or use future information unavailable to the character.
5. Use the original target only to infer the intended stance and response policy. Do not quote, closely paraphrase, or pre-announce its distinctive wording, actions, or thoughts inside <think>.
6. Distinguish stable profile knowledge from episodic scene memories.
7. After </think>, append the original assistant target exactly unchanged.

Return only:

<think>
Natural paragraph 1.

Natural paragraph 2.

Natural paragraph 3.
</think>
{target_assistant_content}
\end{lstlisting}
\end{minipage}
\\
\hline \hline
\end{tabular}
\caption{A reference prompt for LHMT trace generation.}
\label{tab:cot_generation_prompt}
\end{table*}

\begin{table*}[!p]
\centering
\setlength{\tabcolsep}{4pt}
\renewcommand{\arraystretch}{0.95}
\begin{tabular}{p{0.96\textwidth}}
\hline \hline
\rowcolor[gray]{0.9}
\multicolumn{1}{>{\centering\arraybackslash}p{0.96\textwidth}}{\textbf{A Reference Prompt for LHMT trace Validation (Part 1).}} \\
\hline
\begin{minipage}{0.96\textwidth}
\begin{lstlisting}[
basicstyle=\fontsize{8.5pt}{9.2pt}\selectfont\rmfamily,
breaklines=true,
breakatwhitespace=false,
columns=fullflexible,
keepspaces=true,
showstringspaces=false,
frame=none,
aboveskip=2pt,
belowskip=2pt
]
You are a validator for hidden CoT generated for a ShareGPT role-playing memory-activation dataset.

You will receive:
(1) the reference context,
(2) the original assistant target, and
(3) a generated assistant value containing a <think> block followed by the target.

=== REFERENCE CONTEXT ===
{reference_content_with_memories}

=== ORIGINAL ASSISTANT TARGET ===
{target_assistant_content}

=== GENERATED ASSISTANT VALUE ===
{generated_reasoning}

Determine whether the generated <think> is valid supervision for current-turn memory-aware role-playing.

Check the following:

1. Structure
- There is exactly one <think> block containing three natural paragraphs.
- The assistant target after </think> exactly matches the original target.
- The reasoning contains no headings, bullets, numbered steps, or explicit reasoning tags.

2. Grounding and temporal validity
- Every claim is grounded in the profile, current scenario, dialogue prefix, listed memories, character relationships, or original target.
- The reasoning contains no fabricated facts or future information unavailable to the character at the current turn.

3. Memory decision
- Independently determine whether the listed scene memory pool is needed.
- Do not treat character-profile information as scene memory.
- When memory is needed, verify the selected memory ids and explain the behavioral pressure of each memory.
- For multiple activated memories, classify their combined effect as aligned, opposed, or mixed.
- When memory is not needed, verify that the current context and profile are sufficient.

4. Target leakage
- The reasoning may use the target to infer the intended stance and response policy, but must not quote, closely paraphrase, or pre-announce its distinctive wording, actions, thoughts, or withheld information.

5. Training usefulness
- The reasoning clearly explains the current information need, memory relevance, and response policy.
- It is not generic or merely a restatement of the target.
\end{lstlisting}
\end{minipage}
\\
\hline \hline
\end{tabular}
\caption{A reference prompt for LHMT trace validation, part 1.}
\label{tab:cot_validation_prompt_part1}
\end{table*}

\begin{table*}[!p]
\centering
\setlength{\tabcolsep}{4pt}
\renewcommand{\arraystretch}{0.95}
\begin{tabular}{p{0.96\textwidth}}
\hline \hline
\rowcolor[gray]{0.9}
\multicolumn{1}{>{\centering\arraybackslash}p{0.96\textwidth}}{\textbf{A Reference Prompt for LHMT trace Validation (Part 2).}} \\
\hline
\begin{minipage}{0.96\textwidth}
\begin{lstlisting}[
basicstyle=\fontsize{8.5pt}{9.2pt}\selectfont\rmfamily,
breaklines=true,
breakatwhitespace=false,
breakautoindent=false,
breakindent=0pt,
columns=fullflexible,
keepspaces=true,
showstringspaces=false,
frame=none,
aboveskip=2pt,
belowskip=2pt
]
Return ONLY a JSON object, with no preamble or explanation outside JSON:

{
  "pass": true | false,
  "memory_analysis": {
    "need_scene_memory_pool": true | false,
    "memory_ids": ["Memory N", "..."],
    "profile_needed": true | false,
    "current_context_sufficient": true | false,
    "effect_pattern": "none" | "single" | "aligned" | "opposed" | "mixed",
    "memory_effects": [
      {
        "memory_id": "Memory N",
        "direction": "supports_response" | "opposes_response" | "constrains_response" | "complicates_response" | "contextualizes_response",
        "behavioral_pressure": "What this memory pushes the character toward or away from in the current turn."
      }
    ],
    "net_effect": "The combined behavioral effect of the activated memories; empty if no memory is activated."
  },
  "issues": [
    {
      "severity": "critical" | "major" | "minor",
      "type": "structure" | "target_mismatch" | "source_grounding" | "current_turn_boundary" | "future_knowledge_leakage" | "withheld_information_leakage" | "target_wording_leakage" | "memory_decision" | "memory_effect_analysis" | "profile_memory_confusion" | "answer_leakage" | "training_usefulness",
      "message": "A concrete description of the problem."
    }
  ],
  "revision_feedback": "Concise and actionable revision guidance; empty if pass is true."
}

Always provide your best independent memory analysis, even when pass is false.

Severity:
- critical: invalid structure, target mismatch, unsupported or future information, hidden-information leakage, or an incorrect memory-needed decision;
- major: incorrect or incomplete memory use, profile-memory confusion, target leakage, or an unclear response policy;
- minor: limited style or specificity issues that do not change correctness.

If all checks pass, return pass=true, issues=[], and revision_feedback="".
\end{lstlisting}
\end{minipage}
\\
\hline \hline
\end{tabular}
\caption{A reference prompt for LHMT trace validation, part 2.}
\label{tab:cot_validation_prompt_part2}
\end{table*}

\begin{table*}[!p]
\centering
\setlength{\tabcolsep}{4pt}
\renewcommand{\arraystretch}{0.95}
\begin{tabular}{p{0.96\textwidth}}
\hline \hline
\rowcolor[gray]{0.9}
\multicolumn{1}{>{\centering\arraybackslash}p{0.96\textwidth}}{\textbf{A Reference Prompt for Critique Generation in OPSD.}} \\
\hline
\begin{minipage}{0.96\textwidth}
\vspace{0.45em}
\fontsize{8.5pt}{9.2pt}\selectfont
\raggedright

You are a critique expert for long-horizon memory reasoning in role-playing agents. Your task is to identify weaknesses in a model-generated reasoning trajectory by comparing it with a privileged reference, using a structured multi-dimensional rubric.

\vspace{0.9em}
\noindent === INPUTS ===

\vspace{0.25em}
\noindent Current input \(x\):
\begin{itemize}[label=--,leftmargin=1.7em,itemsep=0.02em,topsep=0.08em,parsep=0pt]
    \item Dialogue context \(C_t\): \texttt{\{dialogue\_context\}}
    \item Character profile \(P\): \texttt{\{character\_profile\}}
    \item Memory pool \(M_{r,s}\): \texttt{\{memory\_pool\}}
\end{itemize}

\noindent Model-generated trajectory \(\hat{o}\):
\begin{itemize}[label=--,leftmargin=1.7em,itemsep=0.02em,topsep=0.08em,parsep=0pt]
    \item LHMT trace \(\hat{T}\): \texttt{\{generated\_trace\}}
    \item Role-playing response \(\hat{y}\): \texttt{\{generated\_response\}}
\end{itemize}

\noindent Reference trajectory \(o^*\) (PRIVILEGED -- used for grounding only):
\begin{itemize}[label=--,leftmargin=1.7em,itemsep=0.02em,topsep=0.08em,parsep=0pt]
    \item LHMT trace \(T_t\): \texttt{\{reference\_trace\}}
    \item Role-playing response \(y_t\): \texttt{\{reference\_response\}}
\end{itemize}

\noindent Rubric \(R\): \texttt{\{rubric\_dimensions\_with\_weights\}}\\

\vspace{0.9em}
\noindent === TASK ===

\vspace{0.25em}
Compare \(\hat{o}\) against \(o^*\), guided by rubric \(R\), and identify the rubric dimensions where \(\hat{o}\) meaningfully deviates from \(o^*\). For each deviation, produce a structured diagnosis with:

\begin{enumerate}[leftmargin=1.65em,itemsep=0.03em,topsep=0.10em,parsep=0pt]
    \item \texttt{rubric\_dimension} -- which dimension from \(R\) this deviation belongs to.
    \item \texttt{severity} -- ``high'' \textbar{} ``medium'' \textbar{} ``low'', jointly determined by the rubric weight and the magnitude of deviation.
    \item \texttt{specific\_failure} -- concrete description of what \(\hat{o}\) did wrong (citing \(\hat{o}\)'s content is safe; \(\hat{o}\) is the student's own rollout).
    \item \texttt{improvement\_direction} -- abstract guidance on what the model should do differently, at the level of reasoning pattern, memory usage, or response policy.
    \item \texttt{anchor\_constraints} -- character-level invariants extracted from \(o^*\) that must be preserved (traits, relational facts, decision stance, tone), expressed abstractly.
\end{enumerate}

Cover \(\hat{T}\) (LHMT trace), \(\hat{y}\) (visible response), and their correspondence -- whether \(\hat{y}\) faithfully executes what \(\hat{T}\) plans (memories \(\hat{T}\) activates should be reflected in \(\hat{y}\); response policy \(\hat{T}\) states should be followed by \(\hat{y}\)). Deviations may occur in context understanding, memory activation, response strategy planning, surface response, or trace-response correspondence.

\vspace{0.9em}
\noindent === CRITICAL CONSTRAINTS (surface-leakage prevention) ===

\vspace{0.25em}
The critique will be used as privileged supervision for the student model. The student must not be able to reconstruct \(o^*\)'s exact wording. Therefore:

\begin{itemize}[label=--,leftmargin=1.7em,itemsep=0.03em,topsep=0.10em,parsep=0pt]
    \item DO NOT quote, paraphrase, or reproduce \(o^*\)'s specific surface wording (\(T_t\)'s exact sentences, \(y_t\)'s exact phrases, distinctive word choices).
    \item Refer to \(o^*\) ONLY at the level of: (a) reasoning patterns (``the reference activates Memory 2 and reasons about its denial implication''), (b) decision logic (``the reference adopts a reluctant-denial stance''), (c) character-level invariants (``preserve honest-denial pattern, short reply length''), or (d) rubric-dimension properties.
    \item Quoting \(\hat{o}\) is fine -- it is the student's own rollout.
    \item \texttt{improvement\_direction} must NEVER contain the specific surface form \(y_t\) uses. E.g., ``should explicitly activate Memory 2 to ground a short denial'' -- NOT ``should say `I didn't put my name in the Goblet'''.
\end{itemize}

\vspace{0.9em}
\noindent === OUTPUT FORMAT ===

\vspace{0.25em}
Return ONLY a JSON array. No preamble, no markdown fence, no commentary outside the array. If \(\hat{o}\) has no significant deviations, return \texttt{[]}.

\begin{quote}
\ttfamily\fontsize{8.3pt}{9.1pt}\selectfont
[\\
\quad \{\\
\quad\quad "rubric\_dimension": "<dimension name from R>",\\
\quad\quad "severity": "high | medium | low",\\
\quad\quad "specific\_failure": "<what \(\hat{o}\) did wrong; may cite \(\hat{o}\)'s content>",\\
\quad\quad "improvement\_direction": "<abstract direction; NO \(o^*\) surface wording>",\\
\quad\quad "anchor\_constraints": "<abstract character/relational invariants from \(o^*\)>"\\
\quad \}\\
]
\end{quote}

Order entries by severity (high \(\rightarrow\) low). At most 6 entries; focus on the most consequential deviations.

\vspace{0.45em}
\end{minipage}
\\
\hline \hline
\end{tabular}
\caption{Prompt template for critique generation in OPSD.}
\label{tab:opsd_critique_prompt}
\end{table*}

\end{document}